\documentclass{article}
\usepackage{iclr2027_conference,times}

\usepackage{amsmath,amsfonts,bm}

\def\eqref#1{equation~\ref{#1}}
\def\1{\bm{1}}

\DeclareMathAlphabet{\mathsfit}{\encodingdefault}{\sfdefault}{m}{sl}
\SetMathAlphabet{\mathsfit}{bold}{\encodingdefault}{\sfdefault}{bx}{n}

\usepackage{url}
\usepackage{algorithm}
\usepackage{algpseudocode}
\usepackage{wrapfig}
\usepackage[table]{xcolor}
\usepackage{multirow}
\usepackage{booktabs}
\usepackage{graphicx}
\usepackage{subcaption}
\usepackage{amsmath,amssymb,natbib,float}
\floatstyle{ruled}
\newfloat{trajcfgalgorithm}{tbp}{loa}
\floatname{trajcfgalgorithm}{Algorithm}

\usepackage{hyperref}
\usepackage{cleveref}
\crefformat{equation}{Eq.~(#2#1#3)}
\renewcommand{\eqref}[1]{\textup{(\ref{#1})}}

\definecolor{linkblue}{HTML}{0B5394}
\definecolor{citegreen}{HTML}{1B7F5C}
\definecolor{urlteal}{HTML}{2A6F8F}
\hypersetup{
  colorlinks=true,
  pdfborder={0 0 0},
  linkcolor=linkblue,
  citecolor=citegreen,
  urlcolor=urlteal,
  pdftitle={TrajCFG: Token-Wise Classifier-Free Guidance via Denoising Trajectory Feedback},
  pdfauthor={Jingyu Liu, Hao Jiang, Yikang Ding, Huafeng Shi, Xu Yang, Xiaoqiang Liu, Guoxin Zhang, Pengfei Wan, Kun Gai}
}

\title{TrajCFG: Token-Wise Classifier-Free Guidance\\
via Denoising Trajectory Feedback}

\author{
\textbf{Jingyu Liu}$^{1\dagger\S}$
\quad
\textbf{Hao Jiang}$^{2\dagger}$
\quad
\textbf{Yikang Ding}$^{2\ddagger}$
\quad
\textbf{Huafeng Shi}$^{2}$
\quad
\textbf{Xu Yang}$^{2}$
\\
\textbf{Xiaoqiang Liu}$^{2}$
\quad
\textbf{Guoxin Zhang}$^{2}$
\quad
\textbf{Pengfei Wan}$^{2}$
\quad
\textbf{Kun Gai}$^{2*}$
\\[0.3em]
$^{1}$Renmin University of China
\qquad
$^{2}$KlingAI Research
}

\iclrfinalcopy

\begin{document}

\maketitle
\fancyhead{}
\lhead{Preprint.}
\begingroup
\renewcommand{\thefootnote}{\fnsymbol{footnote}}
\footnotetext[2]{Equal contribution.}
\footnotetext[4]{This work was conducted during the author's internship at KlingAI.}
\footnotetext[3]{Project Leader.}
\footnotetext[1]{Corresponding author.}
\endgroup

\begin{abstract}
Classifier-free guidance (CFG) typically applies a shared scale to all
latent tokens, although their velocity predictions evolve differently
during sampling. We introduce Trajectory-Adaptive Classifier-Free
Guidance (TrajCFG), a training-free method for flow-matching models
that adapts token-wise guidance using historical changes in guided
velocity direction. TrajCFG maps this feedback to bounded scales,
assigning stronger guidance to larger directional changes and weaker
guidance to more consistent directions. It combines adjacent-step
and interval-normalized longer-span feedback, reusing standard CFG
predictions without additional model evaluations. Controlled diagnostics
and ablations examine the roles of directional feedback, token-wise
allocation, and history-span fusion. Experiments on
text-to-video, image-to-video, text-to-audio-video, text-to-image,
and image-to-3D generation show improvements over base samplers
across multiple backbones. For text-to-video generation, TrajCFG
improves VBench Total scores by 0.30--0.75 points over the strongest
evaluated baseline on each of three backbones. These results support
historical directional feedback as a broadly applicable signal for
fine-grained guidance allocation. The project page is available at
\href{https://jyliu-98.github.io/trajcfg-project-page/}%
{https://jyliu-98.github.io/trajcfg-project-page/}.
\end{abstract}

\section{Introduction}
\label{sec:introduction}

Flow matching~\citep{lipman2022flow} provides a common framework
for generative modeling by learning velocity fields that transport
noise into data.
Its applications span image, video, audiovisual, and 3D generation.
Across these tasks, classifier-free guidance
(CFG)~\citep{ho2022classifier} steers generation toward desired
conditions by combining target- and reference-conditioned predictions.
Standard CFG typically applies a shared guidance scale to all
latent sites at a given sampling step.
Such uniform guidance does not account for differences in how
local predictions evolve along the sampling trajectory,
motivating adaptation to local sampling behavior.

Recent work has improved CFG by adapting when and how guidance
is applied.
Guidance schedules vary its strength across noise
levels~\citep{kynkaanniemi2024applying,wang2024analysis}, while
spatially adaptive and velocity-based methods determine the
scale from local guidance energy or within-step velocity
alignment~\citep{li2026delta,luo2026vags}.
Other approaches modify the guided update through optimized
scaling or feedback corrections~\citep{fan2025cfg,wang2026cfg}.
Inference-time refinement also uses a pretrained generator's
own predictions to improve samples through additional
sampling operations~\citep{jang2026self}.
These developments raise a natural question:
\emph{can directional changes in guided velocities already
computed during sampling provide feedback for adapting
CFG at each latent site?}

Our starting point is that guided velocity history records
how the sampler revises its local predictions.
Under Euler sampling, changes in guided velocity correspond
to revisions of the sampler's clean-endpoint estimate.
Separating changes in direction from changes in magnitude
provides a normalized directional statistic.
Video visualizations show that historical velocity angles
can exhibit content-related spatial structure.
A controlled local reconstruction diagnostic further shows
that reducing guidance tends to benefit sites with small
historical angles, whereas increasing guidance tends to
benefit those with larger angles.
Together, these observations motivate guidance allocation
based on historical directional changes.
We evaluate the effectiveness of this principle across
modalities through generation experiments.

\begin{wrapfigure}{r}{0.48\textwidth}
    \centering
    \includegraphics[width=\linewidth]{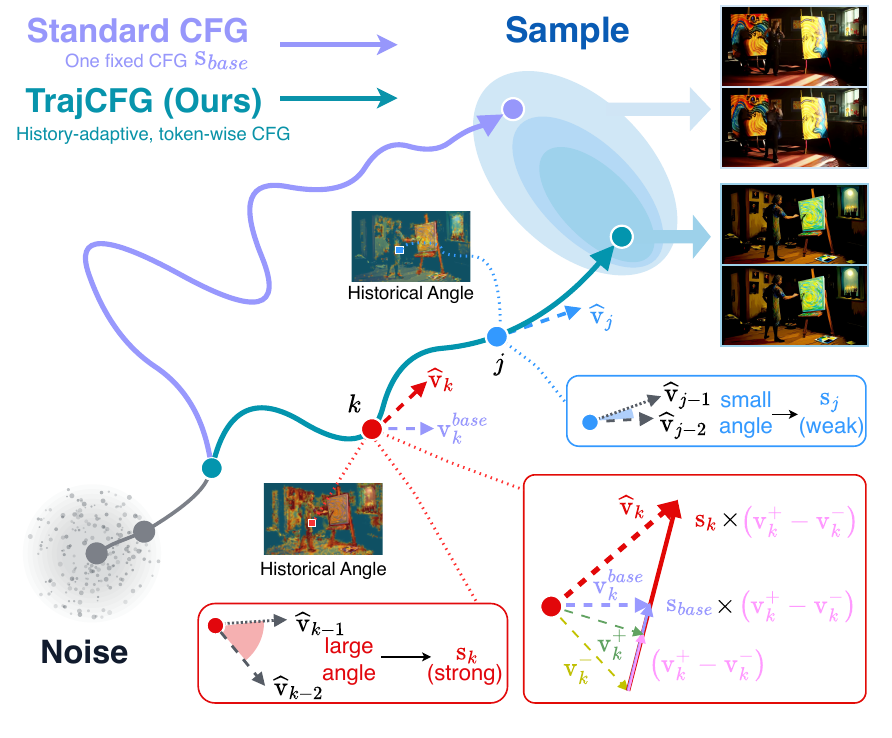}
    \caption{
    \textbf{Overview of TrajCFG.}
    Standard CFG uses a fixed scale (purple), while TrajCFG
    (teal) adapts each token's scale using guided velocities
    from the two most recent completed steps.
    Larger historical angles yield stronger guidance;
    smaller angles yield weaker guidance.
    Predictions are reused without extra model evaluations.
    For clarity, the illustration shows adjacent-step feedback.
    Trajectories are schematic.
    }
    \label{fig:trajcfg_overview}
\end{wrapfigure}

We introduce \textbf{Trajectory-Adaptive Classifier-Free Guidance
(TrajCFG)}, a training-free method for token-wise guidance
adaptation in flow matching models.
As illustrated in Figure~\ref{fig:trajcfg_overview}, TrajCFG
compares guided velocities from the two most recent completed
steps at each latent site and maps their cosine similarity
to a bounded guidance scale.
Larger directional changes receive stronger guidance,
while more consistent directions receive weaker guidance.
We use these scales to combine the current target- and
reference-conditioned predictions.
The guided velocities are then supplied to the solver
and retained for subsequent steps, forming a feedback loop
in which past sampling behavior informs future conditioning
strength.
We also incorporate interval-normalized longer-span feedback
to complement the adjacent-step signal.
Because the rule uses only local guided velocity histories
and existing CFG predictions, it applies across modalities
with different latent representations.
TrajCFG requires neither auxiliary networks nor extra model
evaluations and leaves the pretrained model and sampling
solver unchanged.

We evaluate TrajCFG across text-to-video, image-to-video,
text-to-audio-video, text-to-image, and image-to-3D generation.
Experiments show improvements in generation quality and
conditional alignment across diverse model families.
Controlled diagnostics and ablation studies examine the
roles of token-wise allocation, historical directional
feedback, and the combination of history spans.
Together, these results support TrajCFG as a broadly
applicable approach to adaptive CFG in flow matching models.

\section{Related Work}
\label{sec:related_work}

\paragraph{Scheduling and adaptive classifier-free guidance.}
Classifier-free guidance (CFG) combines target- and
reference-conditioned predictions to control generation~\citep{ho2022classifier}.
Guidance schedules and limited-interval guidance vary conditioning
strength with noise level~\citep{wang2024analysis,kynkaanniemi2024applying},
while Adaptive Guidance and Step AG selectively omit guidance
evaluations~\citep{castillo2025adaptive,zhang2025much}.
$C^2$FG and stage-wise analyses motivate time-dependent scales through
score discrepancies and sampling dynamics,
respectively~\citep{gao2026c,jin2026stage}.
Online-feedback CFG selects scales using trained latent-space
evaluators~\citep{papalampidi2026dynamic}, whereas PathGuide uses
on-policy transport alignment for online scale adaptation or offline
schedule fitting~\citep{nevo2026pathguide}.
TrajCFG extends adaptation to token-wise scales driven by cross-step
changes in guided velocity direction.

\paragraph{Geometry-, spatial-, and history-aware guidance.}
CFG++, APG, and TCFG modify guidance through manifold constraints,
projection and momentum, and score-component filtering,
respectively~\citep{chung2025cfg++,sadat2024eliminating,kwon2025tcfg}.
CFG-Zero combines within-step projection with suppression of
initial updates~\citep{fan2025cfg}, while VAGS adjusts a sample-wise
scale using sampling progress and conditional--unconditional velocity
alignment~\citep{luo2026vags}.
For spatial adaptation, S-CFG uses attention-based semantic
regions~\citep{shen2024rethinking}, whereas SAMG derives point-wise
scales from current-step guidance difference energy~\citep{li2026delta}.
History informs APG's momentum, CFG-Ctrl's discrepancy-based feedback
correction~\citep{wang2026cfg}, and HiGS's additional direction from
current and aggregated past denoised predictions~\citep{sadat2026higs}.
TrajCFG instead measures cross-step angular changes of the
\emph{guided} velocity at each fixed latent site to set bounded
token-wise CFG scales for the next step.
The feedback is computed from guided velocities at completed sampling
steps and modulates the current target--reference prediction difference
without introducing a separate update direction.

\begin{wrapfigure}{r}{0.48\textwidth}
    \centering
    \includegraphics[width=\linewidth]{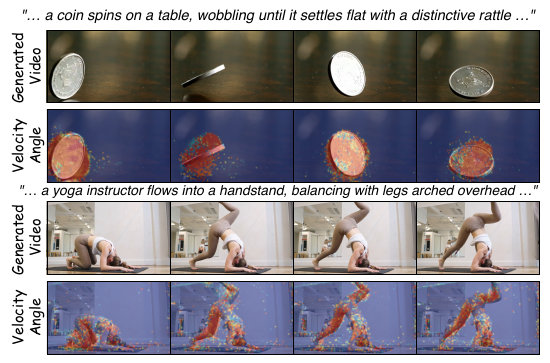}
    \caption{
    \textbf{Content-related structure in historical angles.}
    Generated frames and historical velocity-angle maps at sampling
    step $k=5$ from Wan2.2-14B on Dynamic-Bench. Warmer colors indicate
    larger angular changes.
    }
    \label{fig:velocity_angle_maps}
\end{wrapfigure}

\paragraph{Inference-time refinement and sampling interventions.}
Beyond CFG adaptation, FreeInit refines low-frequency initial
noise~\citep{wu2024freeinit}, while Restart and Zigzag Sampling
introduce additional noising--denoising or inversion
operations~\citep{xu2023restart,bai2025zigzag}.
VideoGuide uses additional teacher or self-guided
steps~\citep{lee2025videoguide}, and Self-Refining Video Sampling
performs fixed-noise predict-and-perturb refinement~\citep{jang2026self}.
FlowMo optimizes latents with a motion objective~\citep{shaulov2026flowmo},
whereas SpecLoR spectrally corrects a clean-latent
lookahead~\citep{zhang2026speclor}.
PAG and STG construct weak predictions through attention perturbation
and layer skipping~\citep{ahn2024self,hyung2025spatiotemporal};
STAS intervenes on sparse massive network
activations~\citep{cheng2026steering}.
TrajCFG uses a short history of velocities already computed during
standard CFG sampling, requiring no auxiliary evaluator, extra
model evaluations, latent optimization, or model-internal changes.
\section{Preliminaries}
\textbf{Flow Matching.}
Flow matching~\citep{lipman2022flow} learns a velocity field
that continuously transforms noise into data.
Under an affine path formulation, the interpolation between
a clean data latent $\mathbf{x}$ and noise $\boldsymbol{\epsilon}$ is
\begin{equation}
    \mathbf{z}_\tau
    = (1-\tau)\mathbf{x} + \tau\boldsymbol{\epsilon},
    \qquad \tau \in [0,1],
    \label{eq:fm_path}
\end{equation}
with path velocity $\boldsymbol{\epsilon}-\mathbf{x}$.
A model $f_\theta(\mathbf{z}_\tau,\tau;y)$ learns the velocity
field conditioned on $y$.
During generation, a numerical solver integrates the learned
field from $\tau=1$ to $\tau=0$, starting from noise.

\textbf{Classifier-Free Guidance.}
At sampling step $k$, given the latent $\mathbf{z}_k$
and time $\tau_k$, a pretrained model predicts velocities
under target conditioning $y^+$ and reference conditioning $y^-$:
\begin{equation}
    \mathbf{v}_k^+ = f_\theta(\mathbf{z}_k,\tau_k;y^+),
    \qquad
    \mathbf{v}_k^- = f_\theta(\mathbf{z}_k,\tau_k;y^-).
    \label{eq:conditional_velocities}
\end{equation}
The reference condition follows the backbone's configuration,
such as a null condition or a negative prompt.
Standard CFG combines these predictions using a baseline guidance
scale $s_{\mathrm{base}}$ shared across latent sites:
\begin{equation}
    \widehat{\mathbf{v}}_k
    = \mathbf{v}_k^-
    + s_{\mathrm{base}}(\mathbf{v}_k^+ - \mathbf{v}_k^-).
    \label{eq:cfg}
\end{equation}
The sampling solver then updates the latent using the guided velocity
$\widehat{\mathbf{v}}_k$.
We use \emph{latent site} to denote a position in the latent
representation and \emph{token} to denote its channel vector;
a token-wise guidance scale is shared across that vector's channels.

\section{Method}
\label{sec:method}
TrajCFG adapts the guidance scale at each latent site using
historical changes in guided velocity direction.
We first motivate this feedback signal, then define the
token-wise allocation rule and its extension to longer history spans.

\subsection{Motivation: Directional Changes as Guidance Feedback}

Standard CFG in Eq.~(\ref{eq:cfg}) applies a shared guidance
scale $s_{\mathrm{base}}$ across all latent sites, even though
their velocity predictions may evolve differently during sampling.
Prior work shows that guidance utility varies across noise
levels~\citep{kynkaanniemi2024applying, wang2024analysis}
and that local prediction statistics can inform adaptive
guidance~\citep{li2026delta}.
These findings motivate us to investigate whether the
directional evolution of guided velocities provides useful
feedback for allocating guidance across latent sites.

Figure~\ref{fig:velocity_angle_maps} illustrates historical
velocity angles in two video examples.
Larger angles concentrate around the foreground subjects,
while much of the background exhibits smaller angles.
This content-related structure suggests that guided velocity
history captures differences in local sampling behavior.
The statistic itself depends only on velocity histories at
individual latent sites and can be computed across modalities.
We next examine what directional changes represent and
whether they inform local guidance benefits.

\noindent\textbf{Interpreting directional changes.}
At a fixed latent site, with its index omitted, the affine
path formulation gives the clean-endpoint estimate
$\mathbf{D}_k
= \mathbf{z}_k - \tau_k\widehat{\mathbf{v}}_k$.
Under an Euler update,
\begin{equation}
\mathbf{D}_{k+1} - \mathbf{D}_k
= -\tau_{k+1}
\bigl(
\widehat{\mathbf{v}}_{k+1}
- \widehat{\mathbf{v}}_k
\bigr),
\label{eq:endpoint_revision}
\end{equation}
so changes in guided velocity correspond to revisions of
the sampler's clean-endpoint estimate, rather than directly
measuring its error against a ground-truth target.
For successive nonzero velocities, let
$n_k = \|\widehat{\mathbf{v}}_k\|_2$ and define their angle
$\theta_k \in [0,\pi]$ by
\[
\cos\theta_k
=
\frac{
\langle
\widehat{\mathbf{v}}_{k+1},
\widehat{\mathbf{v}}_k
\rangle
}{
n_{k+1}n_k
}.
\]
Their squared difference admits the decomposition
\begin{equation}
\|\widehat{\mathbf{v}}_{k+1}
- \widehat{\mathbf{v}}_k\|_2^2
=
(n_{k+1}-n_k)^2
+
2n_{k+1}n_k(1-\cos\theta_k).
\label{eq:velocity_magnitude_direction}
\end{equation}
The two terms separate changes in magnitude from
magnitude-weighted changes in direction.
Normalizing the directional term by $2n_{k+1}n_k$
yields $1-\cos\theta_k$, which increases with the angle
and is invariant to independent positive rescaling of
the two velocities.
We therefore consider this statistic as a measure of
directional revision over the observed sampling interval.
Appendix~\ref{app:trajcfg_motivation} provides the derivation,
the extension to general solvers, and an analysis of
interval dependence.
This interpretation motivates an empirical question:
do sites with different degrees of historical directional
change benefit from different guidance scales?

\begin{wrapfigure}{r}{0.48\textwidth}
    \centering
    \includegraphics[width=\linewidth]
    {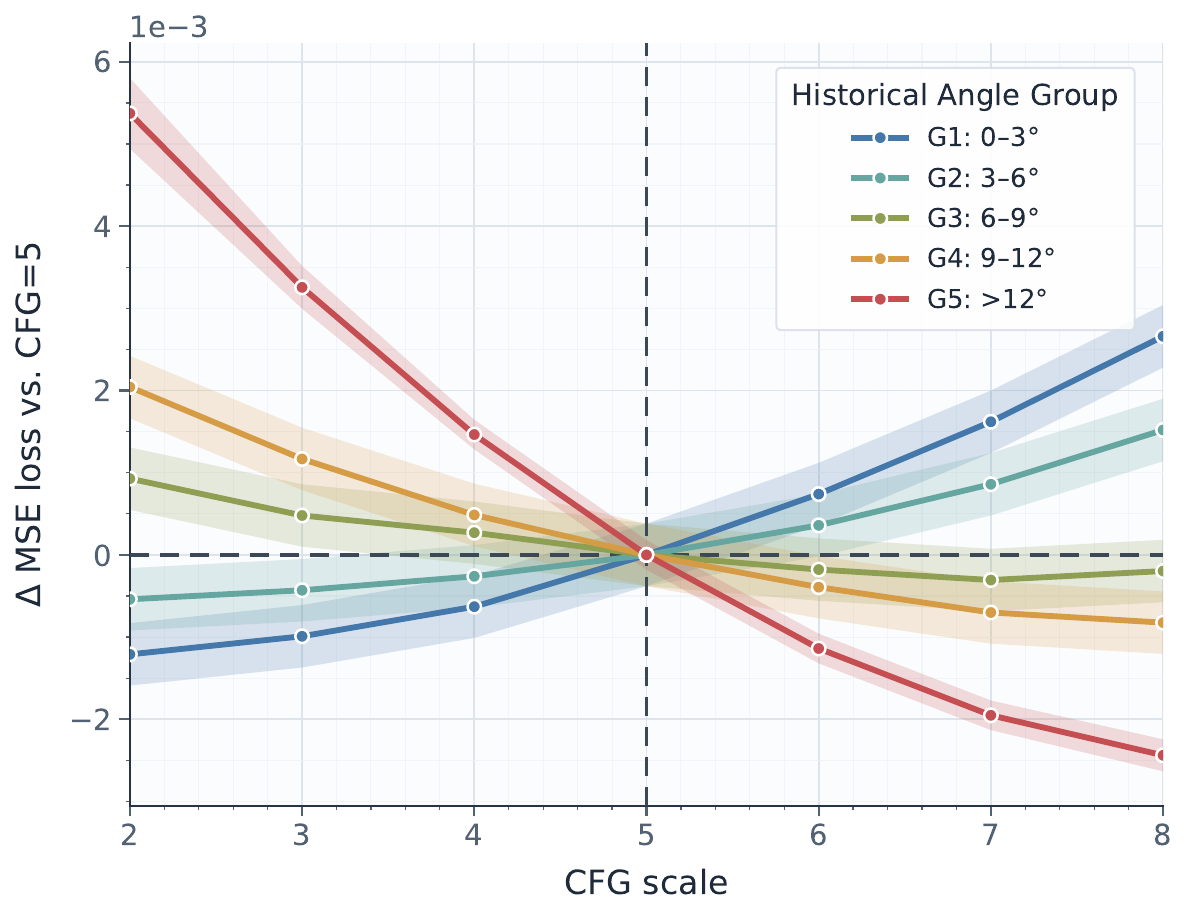}
    \caption{
    \textbf{CFG response by historical angle group.}
    Groups are fixed across candidate CFG scales. Clean-latent MSE
    differences are relative to CFG $=5$ (negative is better);
    shading indicates standard deviations.
    }
    \label{fig:davis_response}
\end{wrapfigure}

\noindent\textbf{Guidance response across directional changes.}
We evaluate Wan2.2-5B on 90 DAVIS~\citep{pont20172017} samples
at all sampling timesteps. For each video, noise seed, and step $k$,
we construct the current and two preceding states by adding noise
to ground-truth clean latents along the affine path, without
sampler rollouts. We group sites by the angle between baseline-guided
velocities at the constructed states for steps $k-2$ and $k-1$.
The five groups in Figure~\ref{fig:davis_response} are
$0$--$3^\circ$, $3$--$6^\circ$, $6$--$9^\circ$,
$9$--$12^\circ$, and $>12^\circ$.
Across candidate CFG scales, we fix these groups, the current
noisy latent, and the target- and reference-conditioned predictions.
We compute channel-averaged clean-endpoint MSE against the
ground-truth latent and aggregate it within each group across
evaluation instances.

Figure~\ref{fig:davis_response} shows the resulting
group-wise MSE differences relative to CFG $=5$.
Among the groups shown, reducing guidance below the
baseline tends to lower error for sites with small
historical angles, whereas increasing guidance tends
to lower error for those with large historical angles.
These controlled comparisons suggest that historical directional
changes carry information about local guidance benefits.

\noindent\textbf{From observation to policy.}
These observations motivate a simple allocation principle:
assign weaker guidance to sites with more consistent
historical directions and stronger guidance to those
with larger directional revisions.
Its use in generation rests on the hypothesis that the
association observed in the controlled reconstruction
diagnostic remains informative along actual sampling
trajectories.
The following subsection implements this principle through
a bounded mapping from historical directional feedback
to token-wise guidance scales.
We evaluate its effectiveness across generative modalities
and examine its components through ablation studies.

\subsection{Token-Wise Guidance from Trajectory Feedback}

Following the allocation principle above, TrajCFG uses
historical directional changes to determine the guidance
scale at each latent site.
Let $\mathbf{z}_k \in \mathbb{R}^{P \times C}$ denote
the latent at sampling step $k$, where $P$ is the number
of sites and $C$ is the number of channels; we omit the
batch dimension.
Each site corresponds to a token indexed by $p$, whose
velocity is a $C$-dimensional vector.
Depending on the backbone, the site axis may represent
spatial, temporal, spatiotemporal, or generic sequence
positions. TrajCFG operates on this flattened site axis
without assuming a modality-specific latent geometry.

\noindent\textbf{Historical directional feedback.}
At step $k \geq 2$, we compare the guided velocities
at each site from the two most recent completed steps:
\begin{equation}
q_{k,p}
=
\operatorname{cos}_{\delta}
\bigl(
\widehat{\mathbf{v}}_{k-2,p},
\widehat{\mathbf{v}}_{k-1,p}
\bigr),
\label{eq:trajcfg_feedback}
\end{equation}
where $\widehat{\mathbf{v}}_{j,p}$ is the guided velocity
actually supplied to the solver at step $j$.
We use the numerically stabilized cosine similarity
\begin{equation}
\operatorname{cos}_{\delta}(\mathbf{a},\mathbf{b})
=
\operatorname{clip}_{[-1,1]}
\left(
\frac{\langle \mathbf{a},\mathbf{b}\rangle}
{
\max(\|\mathbf{a}\|_2,\delta)
\max(\|\mathbf{b}\|_2,\delta)
}
\right),
\label{eq:trajcfg_stabilized_cosine}
\end{equation}
with a norm floor $\delta>0$.
When both velocity norms exceed $\delta$,
$q_{k,p}$ is the cosine of their historical angle:
higher similarity indicates more consistent directions,
while lower similarity indicates larger directional
changes.
At a fixed site, this historical angle corresponds to
$\theta_{k-2}$ in the preceding directional decomposition.
Because both velocities come from completed steps,
the feedback is available before selecting the current
guidance scale.

\noindent\textbf{Mapping feedback to guidance.}
We map historical directional change to guidance strength
using a clipped linear function:
\begin{equation}
g(q)
=
s_{\min}
+
(s_{\max}-s_{\min})
\operatorname{clip}_{[0,1]}
\left(
\frac{1-q}{1-\rho_{\mathrm{lo}}}
\right),
\label{eq:trajcfg_mapping}
\end{equation}
where $0 \leq s_{\min}<s_{\max}$ specifies the guidance
range and $-1 \leq \rho_{\mathrm{lo}}<1$ determines
when the scale reaches $s_{\max}$.
At $q=1$, the scale equals $s_{\min}$.
As similarity decreases, guidance increases linearly
with the directional-change statistic $1-q$,
reaching $s_{\max}$ at $q=\rho_{\mathrm{lo}}$
and remaining capped below this threshold.
This mapping provides a simple bounded realization of
the empirically motivated allocation principle.

\noindent\textbf{Applying token-wise guidance.}
Let $K$ denote the number of sampling updates, indexed
from $0$ to $K-1$, and let
$\mathcal{A} \subseteq \{2,\ldots,K-1\}$
be the set of steps at which adaptation is enabled.
With adjacent-step feedback, the scale applied to token $p$ is
\begin{equation}
s_{k,p}
=
\begin{cases}
g(q_{k,p}), & k \in \mathcal{A},\\
s_{\mathrm{base}}, & k \notin \mathcal{A}.
\end{cases}
\label{eq:trajcfg_scale}
\end{equation}
We then combine the current target- and
reference-conditioned predictions as
\begin{equation}
\widehat{\mathbf{v}}_{k,p}
=
\mathbf{v}^{-}_{k,p}
+
s_{k,p}
\bigl(
\mathbf{v}^{+}_{k,p}
-
\mathbf{v}^{-}_{k,p}
\bigr).
\label{eq:trajcfg_guided_velocity}
\end{equation}
Each scale is shared across the channels of its token.
The resulting guided velocity is passed to the original
solver to update the latent and is retained as history
for subsequent steps.
The first two steps use $s_{\mathrm{base}}$ to
initialize this history.
As sampling proceeds, the feedback therefore reflects
the trajectory produced by the guidance scales
actually applied.

\noindent\textbf{Longer-span feedback.}
We extend the adjacent-step rule with feedback from
more distant historical predictions.
Let $s^{(h)}_{k,p}$ denote the guidance scale obtained
by comparing $\widehat{\mathbf{v}}_{k-1-h,p}$ with
$\widehat{\mathbf{v}}_{k-1,p}$, normalizing the angle
to the adjacent-step interval, and applying the same
guidance mapping.
Appendix~\ref{app:trajcfg_intervals} defines the interval
normalization.
At adaptive steps $k\in\mathcal{A}$, we replace
$g(q_{k,p})$ in Eq.~\eqref{eq:trajcfg_scale} with the
fused guidance scale
\begin{equation}
s_{k,p}
=
\sum_{h \in \mathcal{H}} w_h s^{(h)}_{k,p},
\qquad
w_h \geq 0,\quad \sum_{h \in \mathcal{H}} w_h=1,
\label{eq:trajcfg_fusion}
\end{equation}
where $\mathcal{H}$ contains the available history spans.
We assign decreasing weights to longer spans, prioritizing
recent directional changes that more directly reflect
the current sampling state.
More distant history provides complementary context
with a smaller contribution.
For $k\geq3$, we use adjacent-step ($h=1$) and stride-2
($h=2$) feedback with weights $w_1=0.8$ and $w_2=0.2$.
The additional span compares
$\widehat{\mathbf{v}}_{k-3,p}$ with
$\widehat{\mathbf{v}}_{k-1,p}$;
at $k=2$, we use $\mathcal{H}=\{1\}$ and $w_1=1$.
Steps outside $\mathcal{A}$ retain $s_{\mathrm{base}}$.
This implementation retains three historical velocity
tensors without extra model evaluations.
Algorithm~\ref{alg:trajcfg} in Appendix~\ref{app:trajcfg_algorithm}
summarizes the complete sampling procedure.

\section{Experiments}
\label{sec:experiment}

We compare TrajCFG with inference-time enhancement methods for
text-to-video and image-to-video generation, ablate its guidance
allocation rule, and evaluate three additional generation tasks.

\subsection{Benchmarks}
For T2V, we use VBench~\citep{huang2024vbench} and
Dynamic-Bench~\citep{jang2026self}; for I2V, we use
VBench-I2V~\citep{huang2025vbench++} and
PAI-Bench-G~\citep{zhou2026pai}. We follow the official protocols
for VBench and VBench-I2V. Dynamic-Bench and PAI-Bench-G use the
supported dimensions under the VBench custom evaluation protocol;
these scores are distinct from the official aggregate scores.

\subsection{Backbones and Baselines}
For T2V, we evaluate Wan2.1-1.3B, Wan2.2-5B, and
Wan2.2-14B~\citep{wan2025wan}; for I2V, we use
Cosmos-Predict2.5-2B~\citep{ali2025world} and both Wan2.2 backbones.
We compare with the original sampling configuration (\textit{Base})
and four training-free methods: CFG-Zero~\citep{fan2025cfg},
CFG-Ctrl~\citep{wang2026cfg}, PAG~\citep{ahn2024self}, and
Self-Refining Video Sampling (SRVS)~\citep{jang2026self}.
TrajCFG is labeled \textit{Ours} in figures; guidance hyperparameters
and step counts appear in Appendix Table~\ref{tab:hyperparameters}.

\subsection{Quantitative Comparison}

\begin{table}[!htbp]
\centering
% Runtime is measured; the cross-backbone aggregation rule still needs author documentation.
\caption{
\textbf{Text-to-video generation results on VBench and Dynamic-Bench.}
VBench reports the official Quality (Qual.), Semantic (Sem.), and Total scores.
For Dynamic-Bench, Mot., Over., and Avg. denote Motion Smoothness,
Overall Consistency, and the average over evaluated VBench dimensions under
the custom protocol. Higher scores are better.
Runtime is measured relative to Base ($1.0\times$).
Best and second-best scores in each metric column are bold and underlined,
respectively.
}
\label{tab:t2v-main}
\scriptsize
\setlength{\tabcolsep}{1.8pt}
\resizebox{\columnwidth}{!}{%
\begin{tabular}{l|cccccc|cccccc|cccccc|c}
\toprule
\multirow{3}{*}{\raisebox{-1.2ex}{\textbf{Method}}}
& \multicolumn{6}{c|}{\textbf{Wan2.1-1.3B}}
& \multicolumn{6}{c|}{\textbf{Wan2.2-5B}}
& \multicolumn{6}{c|}{\textbf{Wan2.2-14B}}
& \multirow{3}{*}{\raisebox{-1.2ex}{\textbf{Time}}} \\
\cmidrule(lr){2-7}\cmidrule(lr){8-13}\cmidrule(lr){14-19}
& \multicolumn{3}{c}{\textbf{VBench}}
& \multicolumn{3}{c|}{\textbf{Dynamic-Bench}}
& \multicolumn{3}{c}{\textbf{VBench}}
& \multicolumn{3}{c|}{\textbf{Dynamic-Bench}}
& \multicolumn{3}{c}{\textbf{VBench}}
& \multicolumn{3}{c|}{\textbf{Dynamic-Bench}}
& \\
\cmidrule(lr){2-4}\cmidrule(lr){5-7}
\cmidrule(lr){8-10}\cmidrule(lr){11-13}
\cmidrule(lr){14-16}\cmidrule(lr){17-19}
& \textbf{Qual.} & \textbf{Sem.} & \textbf{Total}
& \textbf{Mot.} & \textbf{Over.} & \textbf{Avg.}
& \textbf{Qual.} & \textbf{Sem.} & \textbf{Total}
& \textbf{Mot.} & \textbf{Over.} & \textbf{Avg.}
& \textbf{Qual.} & \textbf{Sem.} & \textbf{Total}
& \textbf{Mot.} & \textbf{Over.} & \textbf{Avg.}
& \\
\midrule
\textbf{Base}     & 83.03 & 72.77 & 80.98 & 96.40 & 22.36 & 66.18 & 82.88 & 74.52 & 81.21 & 96.58 & 24.00 & 67.45 & 82.93 & 73.05 & 80.96 & 98.14 & 23.47 & 69.14 & 1.0$\times$ \\
\textbf{CFG-Zero} & 82.69 & \underline{73.95} & 80.94 & 97.13 & 23.81 & 66.64 & 83.79 & 74.84 & 82.00 & 96.68 & 24.19 & 67.94 & 83.32 & 75.27 & 81.71 & 98.02 & 25.18 & 69.41 & 1.0$\times$ \\
\textbf{CFG-Ctrl} & 83.65 & 72.68 & 81.45 & 97.28 & \textbf{24.18} & 66.90 & \underline{84.12} & \underline{76.75} & \underline{82.64} & 96.95 & 24.09 & 67.42 & 83.09 & 76.11 & 81.70 & \underline{98.38} & 24.79 & 69.50 & 1.0$\times$ \\
\textbf{PAG}      & 83.24 & 73.92 & 81.38 & 97.15 & 23.89 & 67.25 & 83.37 & 76.14 & 81.93 & 96.91 & 24.11 & 67.63 & 83.07 & 73.31 & 81.12 & 98.11 & 24.90 & 69.46 & 1.5$\times$ \\
\textbf{SRVS}     & \underline{83.69} & 73.10 & \underline{81.57} & \underline{97.38} & 24.15 & \underline{67.48} & 83.93 & 76.63 & 82.47 & 96.87 & \underline{24.30} & \underline{68.11} & \underline{84.24} & \underline{76.59} & \underline{82.71} & 98.26 & \underline{25.24} & \underline{69.86} & 1.5$\times$ \\
\rowcolor{gray!15}
\textbf{TrajCFG}     & \textbf{84.06} & \textbf{75.37} & \textbf{82.32} & \textbf{97.53} & \underline{24.17} & \textbf{67.66} & \textbf{84.82} & \textbf{77.45} & \textbf{83.35} & \textbf{97.02} & \textbf{24.33} & \textbf{68.30} & \textbf{84.41} & \textbf{77.39} & \textbf{83.01} & \textbf{98.43} & \textbf{25.31} & \textbf{70.27} & 1.0$\times$ \\
\bottomrule
\end{tabular}%
}
\label{tab:t2v}
\end{table}

\noindent\textbf{Text-to-video generation.}
Table~\ref{tab:t2v} reports the T2V results.
TrajCFG achieves the highest official VBench Quality, Semantic, and
Total scores on all three backbones, improving Total over the strongest
competitor by 0.30--0.75 points. It also obtains the highest
Dynamic-Bench average and Motion Smoothness on every backbone, with
average-score gains of 0.18--0.41 points over the strongest competitor.

\begin{table}[!htbp]
\centering
% Runtime is measured; the cross-backbone aggregation rule still needs author documentation.
\caption{
\textbf{Image-to-video generation results on VBench-I2V and PAI-Bench-G.}
Img. (Image Alignment) denotes the mean of I2V Subject and I2V Background;
Mot. denotes Motion Smoothness; and Qual. denotes the mean of
Aesthetic Quality and Imaging Quality. Higher scores are better.
Runtime is measured relative to Base ($1.0\times$).
Best and second-best scores in each metric column are bold and underlined,
respectively.
}
\label{tab:i2v-main}
\scriptsize
\setlength{\tabcolsep}{1.8pt}
\resizebox{\columnwidth}{!}{%
\begin{tabular}{l|cccccc|cccccc|cccccc|c}
\toprule
\multirow{3}{*}{\raisebox{-1.2ex}{\textbf{Method}}}
& \multicolumn{6}{c|}{\textbf{Cosmos-Predict2.5-2B}}
& \multicolumn{6}{c|}{\textbf{Wan2.2-5B}}
& \multicolumn{6}{c|}{\textbf{Wan2.2-14B}}
& \multirow{3}{*}{\raisebox{-1.2ex}{\textbf{Time}}} \\
\cmidrule(lr){2-7}\cmidrule(lr){8-13}\cmidrule(lr){14-19}
& \multicolumn{3}{c}{\textbf{VBench-I2V}}
& \multicolumn{3}{c|}{\textbf{PAI-Bench-G}}
& \multicolumn{3}{c}{\textbf{VBench-I2V}}
& \multicolumn{3}{c|}{\textbf{PAI-Bench-G}}
& \multicolumn{3}{c}{\textbf{VBench-I2V}}
& \multicolumn{3}{c|}{\textbf{PAI-Bench-G}}
& \\
\cmidrule(lr){2-4}\cmidrule(lr){5-7}
\cmidrule(lr){8-10}\cmidrule(lr){11-13}
\cmidrule(lr){14-16}\cmidrule(lr){17-19}
& \textbf{Img.} & \textbf{Mot.} & \textbf{Qual.}
& \textbf{Img.} & \textbf{Mot.} & \textbf{Qual.}
& \textbf{Img.} & \textbf{Mot.} & \textbf{Qual.}
& \textbf{Img.} & \textbf{Mot.} & \textbf{Qual.}
& \textbf{Img.} & \textbf{Mot.} & \textbf{Qual.}
& \textbf{Img.} & \textbf{Mot.} & \textbf{Qual.}
& \\
\midrule
\textbf{Base}     & 98.46 & 99.09 & 66.87 & 96.76 & 99.09 & 61.32 & 98.74 & 99.02 & 67.49 & 97.14 & 98.72 & 59.56 & 98.51 & 98.77 & \underline{66.87} & 96.69 & 98.50 & 58.86 & 1.0$\times$ \\
\textbf{CFG-Zero} & \underline{98.53} & \textbf{99.14} & \underline{67.07} & \underline{96.83} & \underline{99.13} & 61.61 & 98.25 & 98.92 & 67.17 & 96.86 & 98.72 & 60.00 & 97.69 & 98.69 & 65.65 & 96.44 & 98.48 & 58.52 & 1.0$\times$ \\
\textbf{CFG-Ctrl} & 98.13 & 98.97 & 66.88 & 95.85 & 98.37 & 60.93 & 98.37 & \underline{99.10} & 66.37 & \underline{97.23} & \underline{98.97} & 59.81 & 98.48 & \underline{98.79} & 66.63 & 96.42 & 98.52 & 58.18 & 1.0$\times$ \\
\textbf{PAG}      & 98.46 & 99.10 & 66.87 & 96.71 & 99.08 & \underline{61.62} & 98.71 & 99.02 & 67.37 & 96.95 & 98.69 & \underline{60.26} & 98.20 & 98.76 & 66.85 & 96.66 & 98.53 & 58.69 & 1.5$\times$ \\
\textbf{SRVS}     & 98.12 & 99.11 & 66.88 & 96.03 & 99.10 & \textbf{61.65} & \underline{98.81} & 99.09 & \underline{67.92} & 96.75 & 98.50 & \textbf{60.39} & \underline{98.57} & \underline{98.79} & 66.84 & \underline{96.73} & \underline{98.61} & \underline{58.86} & 1.5$\times$ \\
\rowcolor{gray!15}
\textbf{TrajCFG}     & \textbf{98.55} & \underline{99.13} & \textbf{67.16} & \textbf{96.89} & \textbf{99.14} & 61.54 & \textbf{98.97} & \textbf{99.32} & \textbf{68.15} & \textbf{97.54} & \textbf{98.99} & 59.69 & \textbf{98.63} & \textbf{98.90} & \textbf{67.02} & \textbf{96.94} & \textbf{98.65} & \textbf{58.90} & 1.0$\times$ \\
\bottomrule
\end{tabular}%
}
\label{tab:i2v}
\end{table}

\noindent\textbf{Image-to-video generation.}
TrajCFG improves every metric summarized in Table~\ref{tab:i2v}
over Base. It achieves the
highest VBench-I2V Image Alignment and Quality on all backbones and
the highest PAI-Bench-G Image Alignment and Motion Smoothness.
PAI-Bench-G Quality gains are smaller and not uniformly best against
all competitors, while the clearest advantages remain reference-image
preservation and motion smoothness.
Complete per-dimension video results are provided in the appendix.

\subsection{Qualitative Comparison}
\label{sec:qualitative_video}

\begin{figure}[!t]
    \centering
    \begin{subfigure}{\linewidth}
        \centering
        \includegraphics[width=\linewidth]{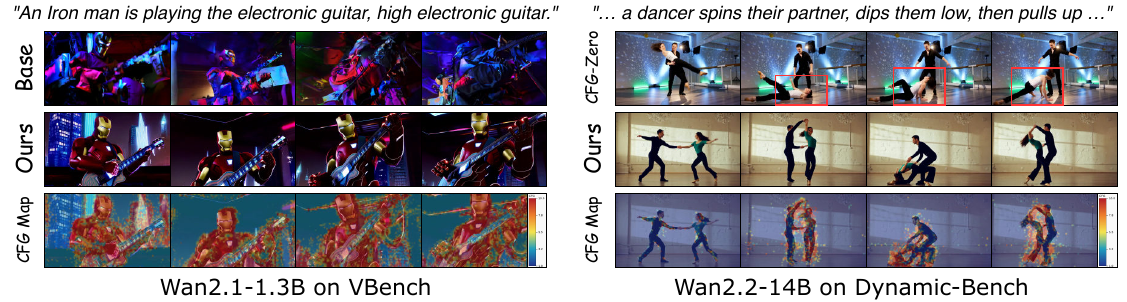}
        \caption{\textbf{Text-to-video (T2V) generation.}}
        \label{fig:t2v_qualitative}
    \end{subfigure}
    \vspace{0.5em}
    \begin{subfigure}{\linewidth}
        \centering
        \includegraphics[width=\linewidth]{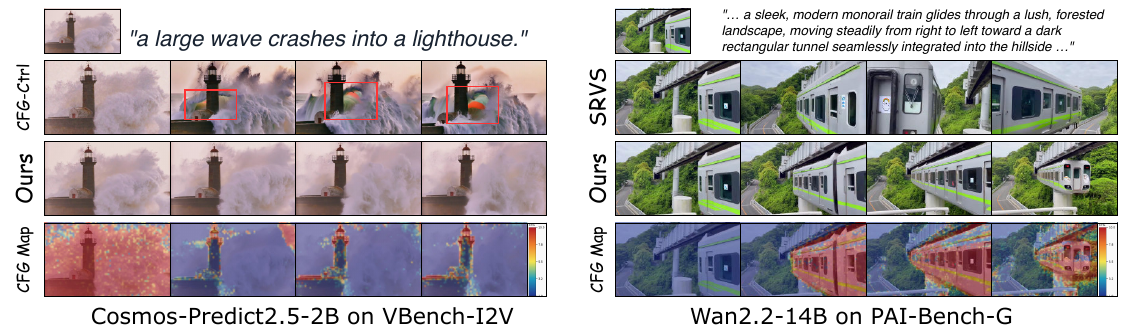}
        \caption{\textbf{Image-to-video (I2V) generation.}}
        \label{fig:i2v_qualitative}
    \end{subfigure}
    \caption{
    \textbf{Qualitative comparisons for video generation.}
    (a) TrajCFG is compared with Base on Wan2.1-1.3B/VBench (left)
    and CFG-Zero on Wan2.2-14B/Dynamic-Bench (right).
    (b) TrajCFG is compared with CFG-Ctrl on
    Cosmos-Predict2.5-2B/VBench-I2V (left) and SRVS on
    Wan2.2-14B/PAI-Bench-G (right); input images appear beside
    the prompts.
    Guidance-scale rows visualize TrajCFG's token-wise scales at
    sampling step $k=5$, with warmer colors indicating stronger
    guidance. Red boxes highlight artifacts in the competing-method
    sequences.
    }
    \label{fig:video_qualitative}
\end{figure}

\noindent\textbf{Text-to-video generation.}
In Figure~\ref{fig:t2v_qualitative}, Base fails to preserve
Iron Man's characteristic appearance, while CFG-Zero abruptly
changes the woman's orientation from front-facing to back-facing
after she contacts the floor. TrajCFG better preserves character
appearance and continuity of body orientation. Its guidance maps
complement the historical angle maps in
Figure~\ref{fig:velocity_angle_maps}.

\noindent\textbf{Image-to-video generation.}
In Figure~\ref{fig:i2v_qualitative}, CFG-Ctrl introduces unnatural
iridescent colors in the seawater, while the SRVS monorail leaves
its track. TrajCFG reduces these color artifacts and keeps the
train aligned with the track. Further T2V/I2V cases appear in
Appendix Figures~\ref{fig:additional_t2v_qualitative}
and~\ref{fig:additional_i2v_qualitative}.

\subsection{Ablation Study}

Table~\ref{tab:ablation} reports ablations of adaptation granularity,
feedback--allocation correspondence, and history-span fusion
on VBench using Wan2.1-1.3B and Wan2.2-5B.

\begin{table}[!htbp]
\centering
\caption{
\textbf{Ablation study of TrajCFG for text-to-video generation on VBench.}
We report the official Quality, Semantic, and Total scores (higher is better).
Fusion weights $(w_1,w_2)$ correspond to adjacent-step ($h=1$) and
stride-2 ($h=2$) guidance estimates, respectively;
TrajCFG (Full) uses $(0.8,0.2)$.
Best and second-best scores in each metric column are bold and underlined,
respectively.
}
\label{tab:ablation}
\small
\setlength{\tabcolsep}{4pt}
\begin{tabular}{ll ccc ccc}
\toprule
& \multirow{2}{*}{\textbf{Variant}}
& \multicolumn{3}{c}{\textbf{Wan2.1-1.3B}}
& \multicolumn{3}{c}{\textbf{Wan2.2-5B}} \\
\cmidrule(lr){3-5}\cmidrule(lr){6-8}
& & \textbf{Quality} & \textbf{Semantic} & \textbf{Total}
& \textbf{Quality} & \textbf{Semantic} & \textbf{Total} \\
\midrule
& TrajCFG (Full)
& \textbf{84.06} & \textbf{75.37} & \textbf{82.32}
& \underline{84.82} & \textbf{77.45} & \textbf{83.35} \\
\midrule
\multirow{2}{*}{\textbf{Q1}}
& Time-only
& \underline{83.94} & 73.17 & 81.78
& 84.62 & \underline{76.94} & 83.08 \\
& Sample-wise
& 83.83 & 74.02 & 81.87
& 84.48 & 76.34 & 82.85 \\
\midrule
\multirow{2}{*}{\textbf{Q2}}
& Shuffled allocation
& 83.77 & 73.67 & 81.75
& 84.43 & 76.73 & 82.89 \\
& Reversed allocation
& 83.70 & 72.96 & 81.55
& 84.55 & 75.82 & 82.81 \\
\midrule
\multirow{2}{*}{\textbf{Q3}}
& Adjacent-only $(1,0)$
& 83.81 & \underline{74.41} & \underline{81.93}
& \textbf{84.85} & 76.93 & \underline{83.27} \\
& Stride-2-only $(0,1)$
& 83.86 & 73.39 & 81.77
& 84.65 & 75.64 & 82.85 \\
\bottomrule
\end{tabular}
\end{table}

\noindent\textbf{Q1: Is token-wise guidance adaptation beneficial?}
We compare TrajCFG with \textit{Time-only}, a fixed schedule averaged
over a calibration set, and \textit{Sample-wise}, which averages each
online scale map across tokens. TrajCFG improves every aggregate score;
its Total-score gains over these variants are 0.45--0.54 points on
Wan2.1-1.3B and 0.27--0.50 points on Wan2.2-5B, supporting adaptation
at individual latent sites.

\noindent\textbf{Q2: Does feedback-aligned guidance allocation matter?}
Using the same initial noise, we replay cached scale maps while either
shuffling token assignments or reversing their strength order; scales
are not recomputed along the altered trajectories. Both variants
underperform the original allocation on every aggregate score,
reducing Total by 0.46--0.77 points. Thus, performance depends on
matching guidance to each token's feedback, not only on the per-step
distribution of scales.

\noindent\textbf{Q3: Does combining history spans improve performance?}
Fusion $(w_1,w_2)=(0.8,0.2)$ outperforms \textit{Adjacent-only}
and \textit{Stride-2-only} in Semantic and Total scores on both
backbones, improving Total over Adjacent-only by 0.39 and 0.08 points.
Adjacent-only has a slightly higher Quality score on Wan2.2-5B.
These results suggest that longer-span feedback modestly complements
the primary adjacent-step signal.

\subsection{Evaluation across Generative Modalities}
\label{sec:cross_modal}

Using each backbone's original configuration as Base, we apply the same
historical feedback rule across three latent representations; qualitative
results appear in Appendix~\ref{app:cross_modal_qualitative},
Figures~\ref{fig:t2av_qualitative}--\ref{fig:i23d_qualitative}.

\noindent\textbf{Text-to-audio-video generation.}
\label{sec:t2av}
We evaluate TrajCFG for text-to-audio-video (T2AV) generation
using LTX-2.3~\citep{hacohen2026ltx} and Ovi~\citep{low2025ovi}
on VABench~\citep{hua2026vabench} and
T2AV-Compass~\citep{cao2025t2av}.
TrajCFG improves all metrics summarized in Table~\ref{tab:t2av}
on both backbones, although some additional metrics decline in the
full evaluation (Appendix~\ref{app:complete_results}).
For Ovi, T2AV-Compass VT increases by 0.485, while VABench
Desync decreases by 0.0579, showing complementary gains in
visual quality and audio--visual synchronization.

\begin{table}[!t]
\centering
\caption{
\textbf{Text-to-audio-video results.}
Audio Aes., T-A, A-V, VT, AA, and DS denote Audio Aesthetics,
text--audio alignment, audio--visual alignment, Video Technological,
Audio Attribute, and DeSync. Lower is better for Desync and DS;
best scores within each backbone are bold.
}
% Metric names: https://nju-link.github.io/T2AV-Compass/
\label{tab:t2av}
\resizebox{\textwidth}{!}{
\begin{tabular}{ll|ccccc|ccccc}
\toprule
\multirow{2}{*}{\textbf{Backbone}} & \multirow{2}{*}{\textbf{Method}} & \multicolumn{5}{c|}{\textbf{VABench}} & \multicolumn{5}{c}{\textbf{T2AV-Compass}} \\
\cmidrule(lr){3-7}\cmidrule(lr){8-12}
& & \textbf{Visual Realism}$\uparrow$ & \textbf{Audio Aes.}$\uparrow$ & \textbf{T-A}$\uparrow$ & \textbf{A-V}$\uparrow$ & \textbf{Desync}$\downarrow$ & \textbf{VT}$\uparrow$ & \textbf{AA}$\uparrow$ & \textbf{T-A}$\uparrow$ & \textbf{A-V}$\uparrow$ & \textbf{DS}$\downarrow$ \\
\midrule
\multirow{2}{*}{\textbf{LTX-2.3}} & \textbf{Base} & 4.615 & 3.345 & 0.3118 & 0.2490 & 0.5911 & 7.295 & 6.709 & 0.1722 & 0.2469 & 0.7140 \\
& \textbf{TrajCFG} & \textbf{4.638} & \textbf{3.351} & \textbf{0.3202} & \textbf{0.2567} & \textbf{0.5685} & \textbf{7.510} & \textbf{6.738} & \textbf{0.1735} & \textbf{0.2481} & \textbf{0.7016} \\
\midrule
\multirow{2}{*}{\textbf{Ovi}} & \textbf{Base} & 4.522 & 3.282 & 0.3288 & 0.1564 & 0.7659 & 6.714 & 6.676 & 0.0897 & 0.1596 & 0.9624 \\
& \textbf{TrajCFG} & \textbf{4.547} & \textbf{3.364} & \textbf{0.3316} & \textbf{0.1654} & \textbf{0.7080} & \textbf{7.199} & \textbf{6.733} & \textbf{0.0922} & \textbf{0.1622} & \textbf{0.9356} \\
\bottomrule
\end{tabular}}
\end{table}

\noindent\textbf{Text-to-image generation.}
\label{sec:t2i}
We evaluate TrajCFG for text-to-image (T2I) generation using
Qwen-Image~\citep{wu2025qwen} and
SD3-Medium~\citep{esser2024scaling} on
GenEval~\citep{ghosh2023geneval} and DPG-Bench~\citep{hu2024ella}.
As shown in Table~\ref{tab:t2i}, TrajCFG improves or matches Base
on every metric, with consistent gains in counting, color, and
attribute binding on GenEval and more modest Overall gains on
DPG-Bench. GenEval Overall increases by 0.0105 and 0.0119 for
Qwen-Image and SD3-Medium, respectively; the corresponding
DPG-Bench Overall gains are 0.08 and 0.43 points.

\begin{table}[!t]
\centering
\caption{\textbf{Text-to-image results.}
Higher is better; best scores within each backbone are bold, including ties.}
\label{tab:t2i}
\resizebox{\textwidth}{!}{
\begin{tabular}{ll|ccccccc|cccccc}
\toprule
\multirow{2}{*}{\textbf{Backbone}} & \multirow{2}{*}{\textbf{Method}} & \multicolumn{7}{c|}{\textbf{GenEval}} & \multicolumn{6}{c}{\textbf{DPG-Bench}} \\
\cmidrule(lr){3-9}\cmidrule(lr){10-15}
& & \textbf{Single Obj.}$\uparrow$ & \textbf{Two Obj.}$\uparrow$ & \textbf{Counting}$\uparrow$ & \textbf{Colors}$\uparrow$ & \textbf{Position}$\uparrow$ & \textbf{Attr. Bind.}$\uparrow$ & \textbf{Overall}$\uparrow$ & \textbf{Global}$\uparrow$ & \textbf{Entity}$\uparrow$ & \textbf{Attribute}$\uparrow$ & \textbf{Relation}$\uparrow$ & \textbf{Other}$\uparrow$ & \textbf{Overall}$\uparrow$ \\
\midrule
\multirow{2}{*}{\textbf{Qwen-Image}} & \textbf{Base} & 0.9875 & 0.9318 & 0.6094 & 0.8324 & \textbf{0.4950} & 0.5825 & 0.7412 & 89.39 & 90.84 & 89.48 & 91.76 & 90.63 & 87.11 \\
& \textbf{TrajCFG} & \textbf{0.9906} & \textbf{0.9394} & \textbf{0.6188} & \textbf{0.8617} & \textbf{0.4950} & \textbf{0.6050} & \textbf{0.7517} & \textbf{91.03} & \textbf{93.20} & \textbf{90.40} & \textbf{93.21} & \textbf{94.35} & \textbf{87.19} \\
\midrule
\multirow{2}{*}{\textbf{SD3-Medium}} & \textbf{Base} & 0.9969 & 0.8611 & 0.6125 & 0.8351 & 0.3125 & 0.5650 & 0.7021 & 86.98 & 84.24 & 86.26 & 87.30 & 85.19 & 83.56 \\
& \textbf{TrajCFG} & \textbf{1.0000} & \textbf{0.8813} & \textbf{0.6344} & \textbf{0.8484} & \textbf{0.3225} & \textbf{0.5975} & \textbf{0.7140} & \textbf{88.35} & \textbf{86.37} & \textbf{86.55} & \textbf{89.38} & \textbf{86.15} & \textbf{83.99} \\
\bottomrule
\end{tabular}}
\end{table}

\noindent\textbf{Image-to-3D generation.}
\label{sec:i23d}
We evaluate TrajCFG for image-to-3D generation using
Hunyuan3D 2.0~\citep{zhao2025hunyuan3d} on
GSO~\citep{downs2022google}.
Table~\ref{tab:i23d} shows improvements over Base on all eight
shape and texture metrics, including lower CMMD and
$\mathrm{FID}_{\mathrm{CLIP}}$. In particular, CMMD and
$\mathrm{FID}_{\mathrm{CLIP}}$ decrease by 0.0039 and 0.0025,
while all four shape similarities also improve. Together, the three evaluations
support applicability across different latent representations.

\begin{table}[!t]
\centering
\caption{
\textbf{Image-to-3D results on GSO.}
Metrics follow Hunyuan3D 2.0; -T and -I denote mesh--text and mesh--image
similarity. Arrows indicate direction; best scores are bold.
}
% Metric suffixes: https://arxiv.org/html/2501.12202v1 (Sec. 5.1).
\label{tab:i23d}
\resizebox{\textwidth}{!}{
\begin{tabular}{ll|cccc|cccc}
\toprule
\multirow{2}{*}{\textbf{Backbone}} & \multirow{2}{*}{\textbf{Method}} & \multicolumn{4}{c|}{\textbf{Shape}} & \multicolumn{4}{c}{\textbf{Texture}} \\
\cmidrule(lr){3-6}\cmidrule(lr){7-10}
& & \textbf{ULIP-T}$\uparrow$ & \textbf{ULIP-I}$\uparrow$ & \textbf{Uni3D-T}$\uparrow$ & \textbf{Uni3D-I}$\uparrow$ & \textbf{CMMD}$\downarrow$ & $\mathbf{FID}_{\mathbf{CLIP}}\downarrow$ & \textbf{CLIP-score}$\uparrow$ & \textbf{LPIPS}$\downarrow$ \\
\midrule
\multirow{2}{*}{\textbf{Hunyuan3D 2.0}} & \textbf{Base} & 0.1753 & 0.2277 & 0.2190 & 0.2714 & 0.0571 & 0.0606 & 0.2503 & 0.1015 \\
& \textbf{TrajCFG} & \textbf{0.1762} & \textbf{0.2290} & \textbf{0.2198} & \textbf{0.2719} & \textbf{0.0532} & \textbf{0.0581} & \textbf{0.2505} & \textbf{0.1008} \\
\bottomrule
\end{tabular}}
\end{table}

\section{Conclusion}

We introduced TrajCFG, a training-free method that adapts token-wise
CFG scales using historical changes in guided velocity direction.
By combining adjacent-step and interval-normalized longer-span
feedback, TrajCFG allocates guidance from sampling history without
additional model evaluations. Experiments show improvements across
video backbones and applicability to audiovisual, image, and 3D
generation. These results support historical directional feedback as a useful
signal for local guidance allocation across latent representations.

Historical directional feedback tracks revisions of the sampler's
predictions rather than directly estimating generation error.
Its effectiveness beyond the evaluated models and sampling
configurations remains to be established, particularly for very
low-step sampling and alternative generative parameterizations.
Future work could investigate self-calibrating guidance rules that
reduce manual parameter selection without additional training or
model evaluations.

\clearpage
\appendix
\section*{Appendix Contents}
\begingroup
\setlength{\parindent}{0pt}
\setlength{\parskip}{0.55em}
\newcommand{\appcontentssection}[2]{%
  \hyperref[#1]{\textbf{\ref*{#1}\quad #2}}\dotfill\pageref*{#1}\par}
\newcommand{\appcontentssubsection}[2]{%
  \hspace*{1.8em}\hyperref[#1]{\ref*{#1}\quad #2}\dotfill\pageref*{#1}\par}
\appcontentssection{app:method_details}{Additional Method Details}
\appcontentssubsection{app:trajcfg_motivation}{Endpoint Revision and Directional Decomposition}
\appcontentssubsection{app:trajcfg_algorithm}{Sampling Algorithm}
\appcontentssection{app:hyperparameters}{Hyperparameter Settings}
\appcontentssection{app:additional_t2v_qualitative}{Additional Text-to-Video Qualitative Comparisons}
\appcontentssection{app:additional_i2v_qualitative}{Additional Image-to-Video Qualitative Comparisons}
\appcontentssection{app:cross_modal_qualitative}{Cross-Modal Qualitative Comparisons}
\appcontentssection{app:complete_results}{Complete Quantitative Results}
\endgroup

\section{Additional Method Details}
\label{app:method_details}

\subsection{Endpoint Revision and Directional Decomposition}
\label{app:trajcfg_motivation}

This section derives the identities used to interpret directional
feedback and explains the interval normalization used for longer
history spans. Unless otherwise stated, all quantities refer to a
fixed latent site, whose index is omitted.

\subsubsection{Endpoint Revision and Solver Dependence}

For the affine path
$\mathbf{z}_{\tau}=(1-\tau)\mathbf{x}+\tau\boldsymbol{\epsilon}$,
the path velocity is $\boldsymbol{\epsilon}-\mathbf{x}$, so
$\mathbf{x}=\mathbf{z}_{\tau}
-\tau(\boldsymbol{\epsilon}-\mathbf{x})$.
Substituting the guided prediction for this velocity defines the
sampler's clean-endpoint estimate:
\begin{equation}
\mathbf{D}_k
=\mathbf{z}_k-\tau_k\widehat{\mathbf{v}}_k.
\label{eq:app_endpoint_estimate}
\end{equation}
Let $\Delta\tau_k=\tau_{k+1}-\tau_k$.
Under the Euler update
$\mathbf{z}_{k+1}=\mathbf{z}_k
+\Delta\tau_k\widehat{\mathbf{v}}_k$,
\begin{equation}
\begin{aligned}
\mathbf{D}_{k+1}-\mathbf{D}_k
&=\mathbf{z}_{k+1}-\mathbf{z}_k
  -\tau_{k+1}\widehat{\mathbf{v}}_{k+1}
  +\tau_k\widehat{\mathbf{v}}_k\\
&=\Delta\tau_k\widehat{\mathbf{v}}_k
  -\tau_{k+1}\widehat{\mathbf{v}}_{k+1}
  +\tau_k\widehat{\mathbf{v}}_k\\
&=-\tau_{k+1}
  \bigl(\widehat{\mathbf{v}}_{k+1}
       -\widehat{\mathbf{v}}_k\bigr).
\end{aligned}
\label{eq:app_endpoint_euler}
\end{equation}
Thus, under Euler sampling, changes in guided velocity correspond
directly to revisions of the clean-endpoint estimate.

For a general solver, define its displacement relative to the Euler
update as
\begin{equation}
\mathbf{e}^{\mathrm{sol}}_k
=\mathbf{z}_{k+1}-\mathbf{z}_k
-\Delta\tau_k\widehat{\mathbf{v}}_k.
\label{eq:app_solver_residual}
\end{equation}
The same calculation gives
\begin{equation}
\mathbf{D}_{k+1}-\mathbf{D}_k
=-\tau_{k+1}
\bigl(\widehat{\mathbf{v}}_{k+1}
     -\widehat{\mathbf{v}}_k\bigr)
+\mathbf{e}^{\mathrm{sol}}_k.
\label{eq:app_endpoint_general}
\end{equation}
This identity also covers multistep solvers, with the effects of
their internal history included in $\mathbf{e}^{\mathrm{sol}}_k$.
The residual is zero for Euler updates and measures departure from
the Euler displacement; it is not a numerical error estimate with
respect to the exact ODE solution.
Both identities concern revisions of the sampler's own endpoint
estimate, rather than its error against a ground-truth target.

\subsubsection{Magnitude--Direction Decomposition}

For successive nonzero guided velocities, define
\begin{equation}
n_k=\|\widehat{\mathbf{v}}_k\|_2,
\qquad
c_k=
\frac{\langle\widehat{\mathbf{v}}_{k+1},
                 \widehat{\mathbf{v}}_k\rangle}
     {n_{k+1}n_k}
=\cos\theta_k.
\label{eq:app_direction_cosine}
\end{equation}
Here $c_k$ compares steps $k$ and $k+1$.
The historical feedback used at step $k$ in the main text instead
compares steps $k-2$ and $k-1$, and equals $c_{k-2}$ when the
norm floor is inactive.
Expanding the squared velocity difference yields
\begin{equation}
\begin{aligned}
\|\widehat{\mathbf{v}}_{k+1}
 -\widehat{\mathbf{v}}_k\|_2^2
&=n_{k+1}^2+n_k^2-2n_{k+1}n_kc_k\\
&=(n_{k+1}-n_k)^2
  +2n_{k+1}n_k(1-c_k).
\end{aligned}
\label{eq:app_velocity_decomposition}
\end{equation}
The first term measures changes in magnitude, and the second
measures magnitude-weighted changes in direction.
Dividing the second term by $2n_{k+1}n_k$ isolates the
directional statistic $1-c_k$.
Equivalently, with unit directions
$\mathbf{u}_k=\widehat{\mathbf{v}}_k/n_k$,
\begin{equation}
1-c_k
=\frac{1}{2}\|\mathbf{u}_{k+1}-\mathbf{u}_k\|_2^2.
\label{eq:app_unit_direction}
\end{equation}
This statistic is invariant to independent positive rescaling of
the velocities. It isolates their directional change, while the
full velocity difference and endpoint revision also depend on
the velocity magnitudes and the sampling time.

\subsubsection{Sampling Intervals and Longer-Span Normalization}
\label{app:trajcfg_intervals}

Suppose the unit direction of a locally nonzero velocity admits
a $C^2$ curve $\mathbf{u}(\tau)$.
For sufficiently small $\Delta>0$, Taylor expansion gives
\begin{equation}
\mathbf{u}(\tau-\Delta)
=\mathbf{u}(\tau)-\Delta\dot{\mathbf{u}}(\tau)
+O(\Delta^2).
\label{eq:app_direction_taylor}
\end{equation}
Using the unit-direction identity above,
\begin{equation}
1-\langle\mathbf{u}(\tau),\mathbf{u}(\tau-\Delta)\rangle
=\frac{\Delta^2}{2}\|\dot{\mathbf{u}}(\tau)\|_2^2
+O(\Delta^3).
\label{eq:app_interval_dependence}
\end{equation}
Directional change therefore depends on both the local directional
rate and the observation interval.
For small angles, $1-\cos\theta\approx\theta^2/2$, motivating
angle rescaling in proportion to the interval length when
comparing nearby history spans.

For span $h$, let
$\Delta_k^{(h)}=|\tau_{k-1}-\tau_{k-1-h}|$ and
$q^{(h)}_{k,p}
=\operatorname{cos}_{\delta}
(\widehat{\mathbf{v}}_{k-1-h,p},
 \widehat{\mathbf{v}}_{k-1,p})$.
We normalize its angle to the adjacent-step interval and apply
the same guidance mapping:
\begin{equation}
\widetilde{q}^{(h)}_{k,p}
=\cos\!\left(
\frac{\Delta_k^{(1)}}{\Delta_k^{(h)}}
\arccos q^{(h)}_{k,p}
\right),
\qquad
s^{(h)}_{k,p}=g(\widetilde{q}^{(h)}_{k,p}).
\label{eq:app_span_normalization}
\end{equation}
For $h=1$, this recovers the adjacent-step feedback.
The rescaling accounts for interval length under a local smoothness
approximation; signals from different spans can still differ
because the direction evolves over their respective intervals.

\noindent\textbf{Scope of interpretation.}
The exact directional identities and rescaling invariance apply
when the velocity norms are nonzero and the numerical norm floor
is inactive. The feedback measures directional change over an
observed interval, rather than step-size-independent curvature.
These relationships explain the signal and its normalization;
the allocation of stronger guidance to larger directional changes
is motivated by the local response diagnostic and evaluated
through generation experiments and ablations.

\subsection{Sampling Algorithm}
\label{app:trajcfg_algorithm}

Algorithm~\ref{alg:trajcfg} presents the sampling procedure
with adjacent-step and longer-span feedback.
The first two steps use the baseline guidance scale
to initialize the velocity history.
At subsequent adaptive steps, guidance is determined
from previously computed guided velocities.
Adjacent-step feedback is used alone at $k=2$;
from $k\geq3$, it is combined with interval-normalized
longer-span feedback using weights $0.8$ and $0.2$.

All cosine similarities are computed token-wise,
and the angle normalization and guidance mapping
are applied element-wise.
The operator $\odot$ broadcasts each token's guidance
scale across its channels.
$\operatorname{SolverStep}$ denotes the original solver
update, including any internal solver state.
We assume a strictly decreasing sampling schedule
$\tau_0>\cdots>\tau_K$.

\begin{algorithm}[tbp]
\caption{TrajCFG with Adjacent-Step and Longer-Span Feedback}
\label{alg:trajcfg}
\begin{algorithmic}[1]
\Require Initial latent $\mathbf{z}_0$;
         schedule $\{\tau_k\}_{k=0}^{K}$
\Require Model $f_\theta$;
         target/reference conditions $y^+,y^-$
\Require Base scale $s_{\mathrm{base}}$;
         adaptive steps
         $\mathcal{A}\subseteq\{2,\ldots,K-1\}$
\Require Guidance mapping $g$ from
         Eq.~\eqref{eq:trajcfg_mapping};
         stabilized cosine $\operatorname{cos}_\delta$
         from Eq.~\eqref{eq:trajcfg_stabilized_cosine}
\Ensure Final latent $\mathbf{z}_K$

\State Initialize an empty velocity cache $\mathcal{V}$

\For{$k=0,\ldots,K-1$}
    \State $\mathbf{s}_k
        \gets s_{\mathrm{base}}\mathbf{1}_P$

    \If{$k\in\mathcal{A}$}
        \State $\mathbf{q}^{\mathrm{short}}
            \gets \operatorname{cos}_\delta
            \bigl(\mathcal{V}[k-2],
                  \mathcal{V}[k-1]\bigr)$
        \State $\mathbf{s}_k
            \gets g(\mathbf{q}^{\mathrm{short}})$

        \If{$k\geq3$}
            \State $\mathbf{q}^{\mathrm{long}}
                \gets \operatorname{cos}_\delta
                \bigl(\mathcal{V}[k-3],
                      \mathcal{V}[k-1]\bigr)$
            \State $\Delta_{\mathrm{short}}
                \gets|\tau_{k-1}-\tau_{k-2}|$
            \State $\Delta_{\mathrm{long}}
                \gets|\tau_{k-1}-\tau_{k-3}|$
            \State $\widetilde{\mathbf{q}}^{\mathrm{long}}
                \gets \cos\!\left(
                \frac{\Delta_{\mathrm{short}}}
                     {\Delta_{\mathrm{long}}}
                \arccos\mathbf{q}^{\mathrm{long}}
                \right)$
            \State $\mathbf{s}_k
                \gets 0.8\,\mathbf{s}_k
                +0.2\,g\bigl(
                \widetilde{\mathbf{q}}^{\mathrm{long}}
                \bigr)$
        \EndIf
    \EndIf

    \State $\mathbf{v}_k^+
        \gets f_\theta(\mathbf{z}_k,\tau_k;y^+)$
    \State $\mathbf{v}_k^-
        \gets f_\theta(\mathbf{z}_k,\tau_k;y^-)$
    \State $\widehat{\mathbf{v}}_k
        \gets \mathbf{v}_k^-
        +\mathbf{s}_k\odot
        (\mathbf{v}_k^+-\mathbf{v}_k^-)$
    \State $\mathbf{z}_{k+1}
        \gets \operatorname{SolverStep}
        (\mathbf{z}_k,\widehat{\mathbf{v}}_k,
         \tau_k,\tau_{k+1})$
    \State $\mathcal{V}[k]
        \gets\widehat{\mathbf{v}}_k$
    \State Retain only the latest three entries
        in $\mathcal{V}$
\EndFor

\State \Return $\mathbf{z}_K$
\end{algorithmic}
\end{algorithm}

The guidance estimates are fused after applying $g$
to each span's feedback.
At non-adaptive steps, the baseline scale is used.
Every step updates the cache with the guided velocity
actually passed to the solver, so subsequent feedback
reflects the realized sampling trajectory.

\clearpage
\begin{samepage}
\section{Hyperparameter Settings}
\label{app:hyperparameters}

Table~\ref{tab:hyperparameters} summarizes the task- and
backbone-specific settings. Here, $s_{\mathrm{base}}$ is the baseline
CFG scale, $s_{\min}$ and $s_{\max}$ define the adaptive guidance range,
$\rho_{\mathrm{lo}}$ is the lower similarity threshold, and the final
column gives the default number of denoising steps.

\begin{table}[H]
\centering
\caption{Hyperparameter settings for each task and backbone.}
\label{tab:hyperparameters}
\small
\setlength{\tabcolsep}{7pt}
\begin{tabular}{llccccc}
\toprule
\textbf{Task} & \textbf{Backbone}
& $\boldsymbol{s_{\mathrm{base}}}$
& $\boldsymbol{s_{\min}}$
& $\boldsymbol{s_{\max}}$
& $\boldsymbol{\rho_{\mathrm{lo}}}$
& \textbf{Steps} \\
\midrule
\multirow{3}{*}{T2V}
& Wan2.1-1.3B & 6.0 & 5.0 & 10.0 & 0.95 & 40 \\
& Wan2.2-5B   & 5.0 & 3.0 & 10.0 & 0.95 & 40 \\
& Wan2.2-14B  & 3.5 & 1.0 & 10.0 & 0.90 & 40 \\
\midrule
\multirow{3}{*}{I2V}
& Cosmos-Predict2.5-2B & 5.0 & 4.0 & 9.0 & 0.90 & 35 \\
& Wan2.2-5B             & 5.0 & 2.0 & 10.0 & 0.95 & 40 \\
& Wan2.2-14B            & 2.5 & 1.0 & 7.0 & 0.95 & 40 \\
\midrule
\multirow{2}{*}{T2AV}
& LTX-2.3 & 5.0 & 4.0 & 9.0 & 0.90 & 30 \\
& Ovi     & 3.5 & 3.5 & 10.0 & 0.95 & 50 \\
\midrule
\multirow{2}{*}{T2I}
& Qwen-Image & 5.0 & 4.0 & 9.0 & 0.95 & 30 \\
& SD3-Medium & 7.0 & 6.0 & 9.0 & 0.95 & 28 \\
\midrule
\multirow{2}{*}{I2-3D}
& Hunyuan3D 2.0 (Shape)   & 7.0 & 5.0 & 9.0 & 0.95 & 50 \\
& Hunyuan3D 2.0 (Texture) & 2.0 & 2.0 & 3.0 & 0.95 & 50 \\
\bottomrule
\end{tabular}
\end{table}
\end{samepage}

\section{Additional Text-to-Video Qualitative Comparisons}
\label{app:additional_t2v_qualitative}

\begin{figure}[!t]
    \centering
    \includegraphics[width=\linewidth]{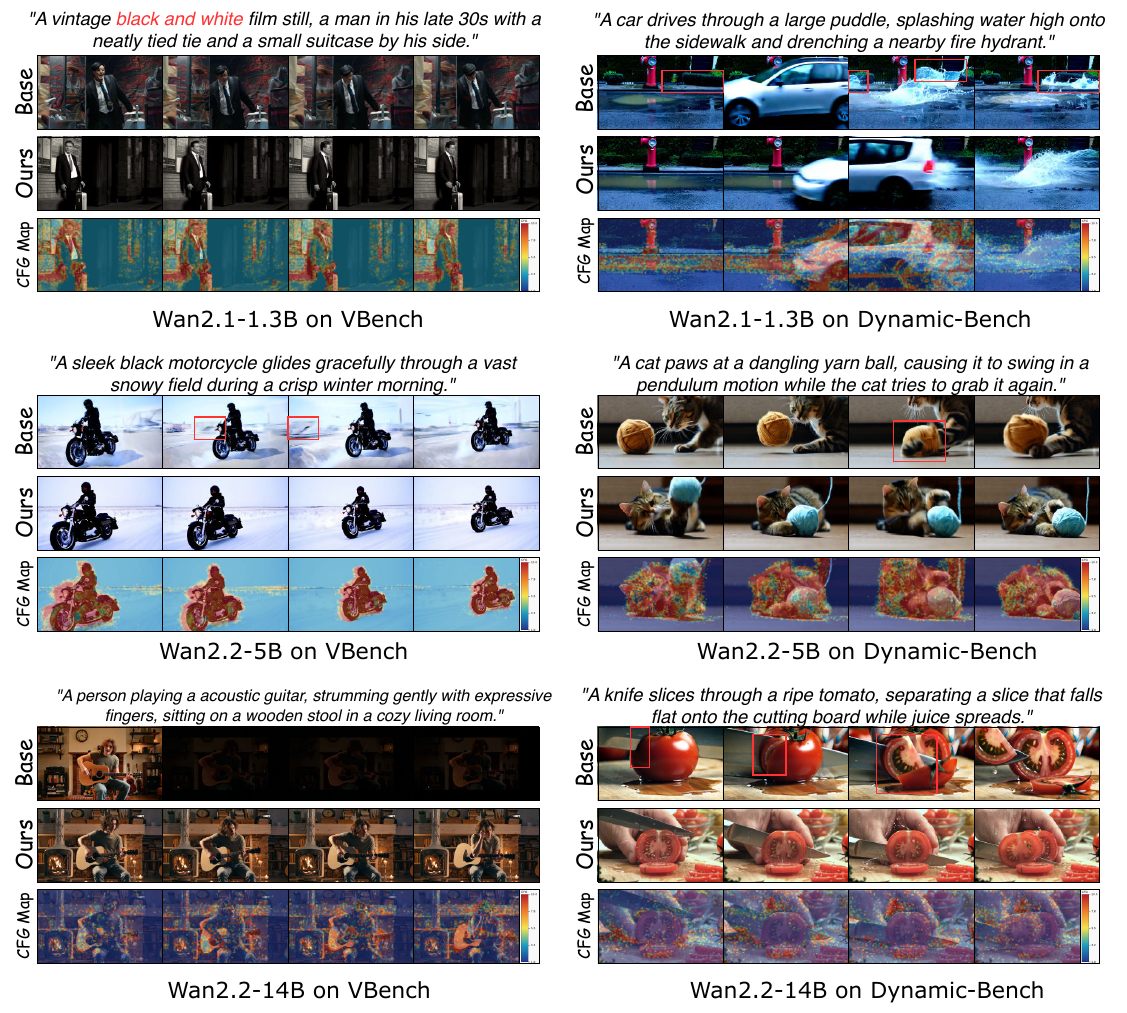}
    \caption{
    \textbf{Additional qualitative comparisons for text-to-video generation.}
    Results are shown for Wan2.1-1.3B, Wan2.2-5B, and Wan2.2-14B on
    VBench and Dynamic-Bench. Each example presents selected frames from
    Base and TrajCFG together with the token-wise guidance map produced
    by TrajCFG; red boxes mark representative Base failures.
    }
    \label{fig:additional_t2v_qualitative}
\end{figure}

Figure~\ref{fig:additional_t2v_qualitative} provides six additional
comparisons spanning appearance, physical consistency, temporal
stability, and action semantics. From upper left to lower right, Base
retains color despite a black-and-white prompt; produces water splashes
in distant dry regions; fails to maintain the snowy setting and depicts
the motorcycle moving backward; causes the cat's paw to intersect the
yarn ball; abruptly darkens the guitar scene; and turns a single slicing
action into two tomato slices. TrajCFG better follows the requested
appearance while improving spatial localization, object interaction,
temporal consistency, and action--result correspondence.

\section{Additional Image-to-Video Qualitative Comparisons}
\label{app:additional_i2v_qualitative}

\begin{figure}[!t]
    \centering
    \includegraphics[width=\linewidth]{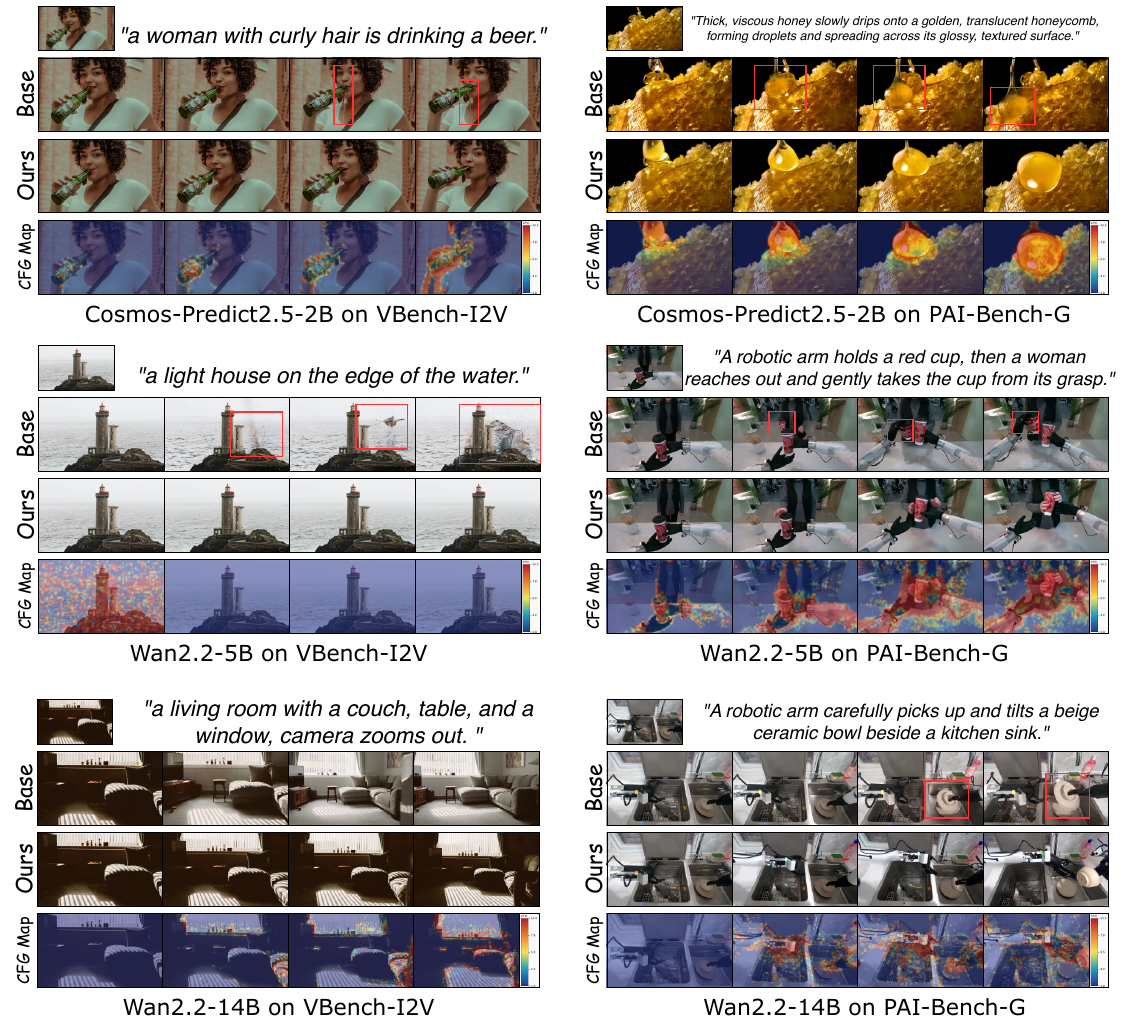}
    \caption{
    \textbf{Additional qualitative comparisons for image-to-video generation.}
    The three rows show Cosmos-Predict2.5-2B, Wan2.2-5B, and Wan2.2-14B,
    while the left and right columns use VBench-I2V and PAI-Bench-G,
    respectively. Each example presents the input image, selected frames
    from Base and TrajCFG, and the token-wise guidance map produced by
    TrajCFG; red boxes mark representative Base failures.
    }
    \label{fig:additional_i2v_qualitative}
\end{figure}

Figure~\ref{fig:additional_i2v_qualitative} provides six comparisons of
physical interaction, material dynamics, structural stability, object
continuity, camera control, and manipulation. From upper left to lower
right, Base lets the beer spill past the woman's mouth instead of being
drunk; makes the honey behave like a solid blob rather than a viscous
liquid that spreads; introduces artifacts and structural distortions;
causes a cup to appear spontaneously in the woman's hand; fails to
execute the requested zoom-out; and lifts the supporting dish together
with the bowl. TrajCFG yields more coherent liquid behavior, geometry,
object transfer, camera motion, and fine-grained physical interactions.

\section{Cross-Modal Qualitative Comparisons}
\label{app:cross_modal_qualitative}

\begin{figure}[!t]
    \centering
    \includegraphics[width=\linewidth]{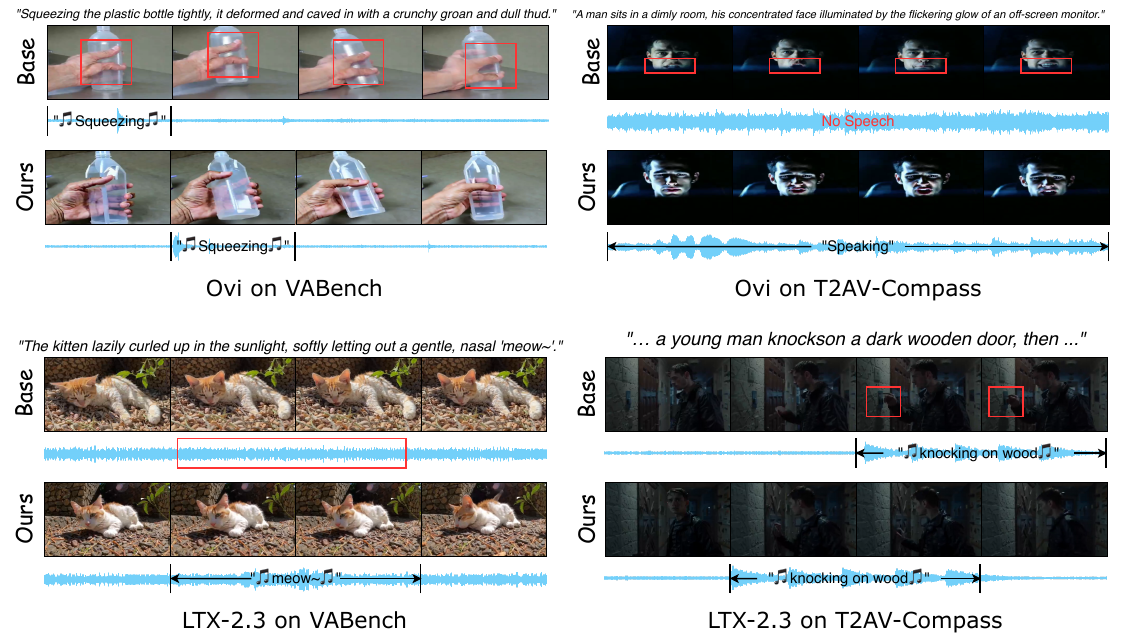}
    \caption{
    \textbf{Qualitative comparisons for text-to-audio-video generation.}
    The four examples show Ovi on VABench (upper left), Ovi on
    T2AV-Compass (upper right), LTX-2.3 on VABench (lower left), and
    LTX-2.3 on T2AV-Compass (lower right). Each example pairs selected
    video frames with the corresponding audio waveform. Arrows mark
    sound-event intervals, while red boxes highlight spatial or
    temporal discrepancies in Base.
    }
    \label{fig:t2av_qualitative}
\end{figure}

Figure~\ref{fig:t2av_qualitative} illustrates complementary visual and
audio--visual failures in Base. In the upper-left example, squeezing the
bottle produces visibly distorted fingers. In the upper-right example,
the subject moves his mouth, yet the accompanying audio contains no
speech. In the lower-left example, Base omits the prompted kitten meow.
Finally, in the lower-right example, Base depicts the hand contacting a
doorbell while generating the sound of knocking on wood, creating a
cross-modal content mismatch. TrajCFG reduces the hand artifact and more
faithfully aligns the visible actions and prompted events with the
generated audio.

\begin{figure}[!t]
    \centering
    \includegraphics[width=\linewidth]{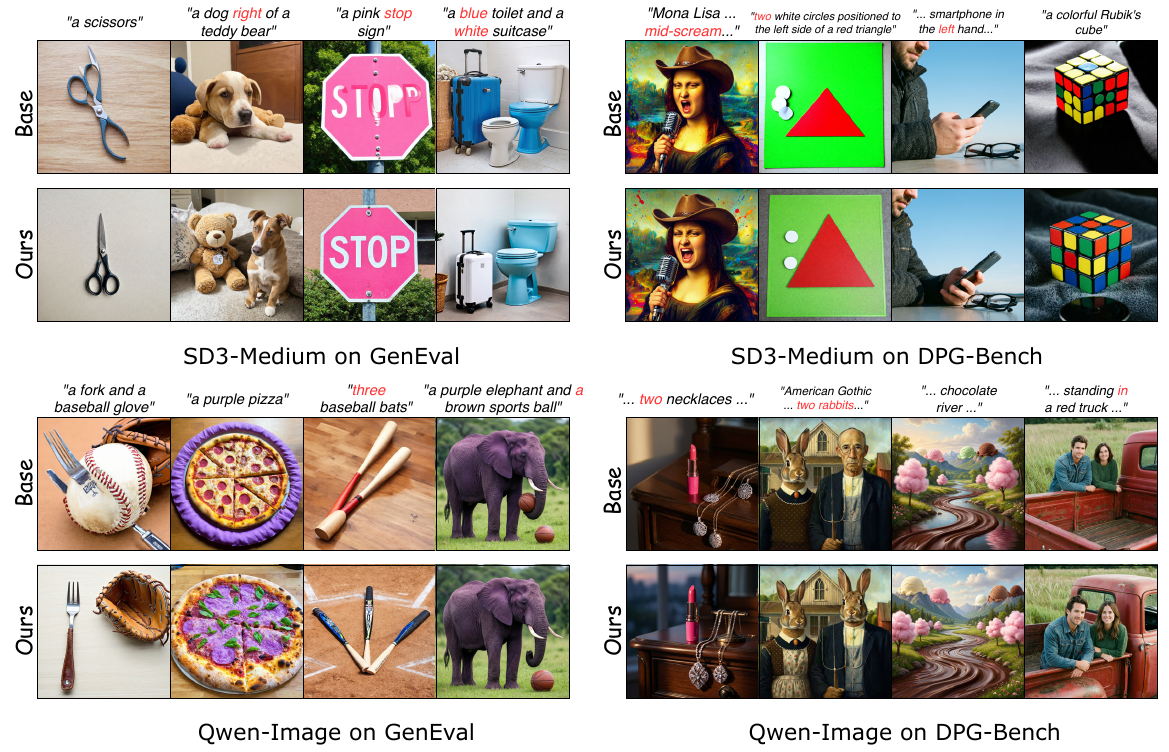}
    \caption{
    \textbf{Qualitative comparisons for text-to-image generation.}
    The four groups show SD3-Medium on GenEval (upper left), SD3-Medium
    on DPG-Bench (upper right), Qwen-Image on GenEval (lower left), and
    Qwen-Image on DPG-Bench (lower right). Within each group, Base and
    TrajCFG are shown in the top and bottom rows, respectively. Red text
    highlights selected prompt requirements.
    }
    \label{fig:t2i_qualitative}
\end{figure}

Figure~\ref{fig:t2i_qualitative} illustrates improvements in prompt
adherence across both backbones. Representative Base failures include
an implausible pair of scissors, an incorrect dog--teddy-bear relation,
swapped colors between the toilet and suitcase, three circles instead
of two, an extra baseball, two bats instead of three, and a person in place
of one of the two requested rabbits. TrajCFG better preserves the
requested object identities, counts, spatial relations, colors, and
attribute bindings while maintaining comparable visual quality.

\begin{figure}[!t]
    \centering
    \includegraphics[width=0.65\linewidth]{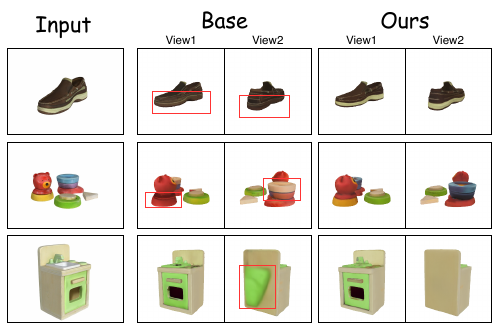}
    \caption{
    \textbf{Qualitative comparisons for image-to-3D generation.}
    Input images and two rendered views from Base and TrajCFG.
    Red boxes highlight appearance discrepancies in Base.
    }
    \label{fig:i23d_qualitative}
\end{figure}

Figure~\ref{fig:i23d_qualitative} shows that TrajCFG better preserves
the light-colored shoe sole and stacked-piece colors, while reducing
apparent texture leakage from the toy kitchen's green front panel.

\section{Complete Quantitative Results}
\label{app:complete_results}

This section reports the complete per-dimension results underlying the summarized tables in the main paper. Tables~\ref{tab:app-full-vbench-original-experiments} and~\ref{tab:app-full-dynamic-bench} present text-to-video (T2V) results on VBench and Dynamic-Bench, respectively, while Table~\ref{tab:app-full-vbench-t2v-ablation} reports the complete VBench ablation results. Tables~\ref{tab:app-full-vbench-i2v} and~\ref{tab:app-full-pai-bench-g} present image-to-video (I2V) results on VBench-I2V and PAI-Bench-G, respectively. Tables~\ref{tab:app-full-vabench-t2av} and~\ref{tab:app-full-t2av-compass} present text-to-audio-video (T2AV) results on VABench and T2AV-Compass, respectively. Unless otherwise specified, higher values indicate better performance. For the T2V and I2V tables, the best and second-best results within each backbone are \textbf{boldfaced} and \underline{underlined}, respectively; TrajCFG rows and the full TrajCFG ablation rows are shaded in gray. For the T2AV tables, only the best results are shown in \textbf{bold}.

\begin{table*}[t]
\centering
\caption{Complete per-dimension results for text-to-video (T2V) generation on VBench.}
\label{tab:app-full-vbench-original-experiments}

\resizebox{\textwidth}{!}{%
\begin{tabular}{llcccccccccc}
\toprule
\multirow{2}{*}{\textbf{Backbone}}
& \multirow{2}{*}{\textbf{Method}}
& \textbf{Aesthetic} & \textbf{Appearance} & \textbf{Background}
& \multirow{2}{*}{\textbf{Color}}
& \textbf{Dynamic} & \textbf{Human} & \textbf{Imaging} & \textbf{Motion} & \textbf{Multiple} & \textbf{Object} \\
& & \textbf{Quality} & \textbf{Style} & \textbf{Consistency} & & \textbf{Degree} & \textbf{Action} & \textbf{Quality} & \textbf{Smoothness} & \textbf{Objects} & \textbf{Class} \\
\midrule

\multirow{6}{*}{\textbf{Wan2.1-1.3B}}
& \textbf{Base}     & 64.11 & 22.04 & 96.70 & 84.71 & 65.28 & 91.00 & 62.92 & 98.00 & 71.19 & 87.42 \\
& \textbf{CFG-Zero} & \underline{65.26} & 22.16 & 95.61 & \textbf{87.67} & \underline{68.90} & \underline{93.00} & \underline{64.91} & 97.41 & 69.74 & 86.31 \\
& \textbf{CFG-Ctrl} & 64.49 & 22.12 & \underline{96.77} & 86.69 & 68.06 & \underline{93.00} & 64.80 & 98.09 & 68.90 & 84.97 \\
& \textbf{PAG}      & 64.76 & 22.24 & 96.58 & 85.62 & 66.67 & \underline{93.00} & 64.71 & 97.89 & 70.50 & 85.76 \\
& \textbf{SRVS}     & 64.87 & \underline{22.37} & 96.64 & 82.77 & 66.67 & 92.00 & \textbf{64.94} & \textbf{98.19} & \underline{71.49} & \underline{88.92} \\
\rowcolor{gray!15}
& \textbf{TrajCFG}  & \textbf{65.48} & \textbf{22.45} & \textbf{96.90} & \underline{86.82} & \textbf{70.83} & \textbf{93.75} & 64.71 & \underline{98.12} & \textbf{75.68} & \textbf{89.14} \\
\midrule

\multirow{6}{*}{\textbf{Wan2.2-5B}}
& \textbf{Base}     & 65.17 & 22.34 & 96.23 & \underline{90.05} & 55.56 & \textbf{95.00} & 63.89 & \underline{98.44} & 65.78 & 79.17 \\
& \textbf{CFG-Zero} & 66.17 & \underline{22.40} & 96.98 & 87.20 & \underline{62.50} & 93.00 & 66.60 & 98.38 & 67.61 & 86.71 \\
& \textbf{CFG-Ctrl} & \underline{66.20} & 22.38 & 97.35 & 88.38 & \underline{62.50} & \underline{94.00} & 65.99 & 98.33 & 71.80 & 87.18 \\
& \textbf{PAG}      & 65.49 & 22.22 & 97.09 & 86.15 & 56.94 & \textbf{95.00} & 66.74 & 98.38 & 71.42 & \textbf{88.61} \\
& \textbf{SRVS}     & \textbf{66.22} & 21.84 & \underline{97.41} & 88.86 & 56.94 & \underline{94.00} & \underline{66.94} & \underline{98.44} & \underline{72.26} & \underline{88.21} \\
\rowcolor{gray!15}
& \textbf{TrajCFG}  & 65.49 & \textbf{22.43} & \textbf{97.53} & \textbf{91.36} & \textbf{68.06} & \textbf{95.00} & \textbf{67.26} & \textbf{98.45} & \textbf{73.86} & 85.68 \\
\midrule

\multirow{6}{*}{\textbf{Wan2.2-14B}}
& \textbf{Base}     & 66.37 & 21.56 & 95.26 & 77.83 & 72.22 & 90.00 & 66.75 & 97.31 & 76.30 & 87.66 \\
& \textbf{CFG-Zero} & 67.12 & 21.78 & 95.09 & \underline{82.04} & 69.44 & 92.00 & 67.36 & 97.69 & 74.54 & 86.08 \\
& \textbf{CFG-Ctrl} & 66.58 & \underline{22.61} & 95.37 & 79.83 & 65.28 & 94.00 & 66.91 & 97.67 & \textbf{78.58} & 88.29 \\
& \textbf{PAG}      & 66.97 & 20.92 & 95.39 & 81.40 & 69.44 & 94.00 & 66.37 & 97.28 & 68.29 & 88.45 \\
& \textbf{SRVS}     & \underline{67.21} & 22.40 & \underline{95.42} & 80.34 & \underline{77.78} & \textbf{95.24} & \textbf{67.70} & \underline{97.71} & 77.52 & \underline{89.64} \\
\rowcolor{gray!15}
& \textbf{TrajCFG}  & \textbf{67.26} & \textbf{22.66} & \textbf{95.43} & \textbf{83.73} & \textbf{79.17} & \underline{95.00} & \underline{67.57} & \textbf{97.73} & \underline{78.20} & \textbf{90.03} \\
\bottomrule
\end{tabular}%
}

\vspace{0.5em}

\resizebox{\textwidth}{!}{%
\begin{tabular}{llcccccc|cc|c}
\toprule
\multirow{2}{*}{\textbf{Backbone}}
& \multirow{2}{*}{\textbf{Method}}
& \textbf{Overall}
& \multirow{2}{*}{\textbf{Scene}}
& \textbf{Spatial} & \textbf{Subject} & \textbf{Temporal} & \textbf{Temporal} & \textbf{Quality} & \textbf{Semantic}
& \multirow{2}{*}{\textbf{Total}} \\
& & \textbf{Consistency} & & \textbf{Relationship} & \textbf{Consistency} & \textbf{Flickering} & \textbf{Style} & \textbf{Score} & \textbf{Score} & \\
\midrule

\multirow{6}{*}{\textbf{Wan2.1-1.3B}}
& \textbf{Base}     & 25.30 & 38.52 & 65.42 & 93.08 & \underline{99.47} & 22.48 & 83.03 & 72.77 & 80.98 \\
& \textbf{CFG-Zero} & 25.35 & 39.10 & 71.52 & 91.01 & 99.03 & 22.77 & 82.69 & \underline{73.95} & 80.94 \\
& \textbf{CFG-Ctrl} & 25.30 & 37.79 & 66.02 & 93.10 & 99.46 & 22.43 & 83.65 & 72.68 & 81.45 \\
& \textbf{PAG}      & \underline{25.49} & 38.84 & \underline{72.58} & 92.39 & 99.33 & \underline{22.83} & 83.24 & 73.92 & 81.38 \\
& \textbf{SRVS}     & 25.48 & \underline{39.17} & 64.97 & \textbf{93.28} & \textbf{99.49} & 22.51 & \underline{83.69} & 73.10 & \underline{81.57} \\
\rowcolor{gray!15}
& \textbf{TrajCFG}  & \textbf{25.61} & \textbf{39.64} & \textbf{72.79} & \underline{93.13} & \textbf{99.49} & \textbf{22.93} & \textbf{84.06} & \textbf{75.37} & \textbf{82.32} \\
\midrule

\multirow{6}{*}{\textbf{Wan2.2-5B}}
& \textbf{Base}     & 24.82 & 43.60 & 77.12 & 93.86 & 99.50 & \underline{23.35} & 82.88 & 74.52 & 81.21 \\
& \textbf{CFG-Zero} & 25.65 & 40.26 & \textbf{77.24} & 92.38 & 99.38 & 23.29 & 83.79 & 74.84 & 82.00 \\
& \textbf{CFG-Ctrl} & 25.75 & \underline{48.69} & 76.98 & \underline{94.12} & \underline{99.51} & 23.34 & \underline{84.12} & \underline{76.75} & \underline{82.64} \\
& \textbf{PAG}      & 24.91 & 48.18 & 75.19 & 92.53 & 99.48 & 23.33 & 83.37 & 76.14 & 81.93 \\
& \textbf{SRVS}     & \underline{25.78} & 47.82 & 77.06 & \textbf{94.46} & 99.40 & 23.25 & 83.93 & 76.63 & 82.47 \\
\rowcolor{gray!15}
& \textbf{TrajCFG}  & \textbf{25.84} & \textbf{49.49} & \underline{77.19} & \textbf{94.46} & \textbf{99.57} & \textbf{23.41} & \textbf{84.82} & \textbf{77.45} & \textbf{83.35} \\
\midrule

\multirow{6}{*}{\textbf{Wan2.2-14B}}
& \textbf{Base}     & 24.15 & 37.94 & 75.69 & 89.42 & 98.88 & 22.59 & 82.93 & 73.05 & 80.96 \\
& \textbf{CFG-Zero} & 25.88 & 45.64 & \underline{77.73} & 91.01 & 98.74 & 22.65 & 83.32 & 75.27 & 81.71 \\
& \textbf{CFG-Ctrl} & \underline{26.10} & 44.58 & 75.69 & \textbf{91.50} & 99.01 & \textbf{23.15} & 83.09 & 76.11 & 81.70 \\
& \textbf{PAG}      & 25.67 & 42.95 & 72.78 & 91.18 & 98.87 & 21.46 & 83.07 & 73.31 & 81.12 \\
& \textbf{SRVS}     & 25.98 & \underline{46.22} & 77.50 & 91.04 & \underline{99.05} & \underline{22.97} & \underline{84.24} & \underline{76.59} & \underline{82.71} \\
\rowcolor{gray!15}
& \textbf{TrajCFG}  & \textbf{26.21} & \textbf{46.95} & \textbf{78.44} & \underline{91.32} & \textbf{99.09} & 22.82 & \textbf{84.41} & \textbf{77.39} & \textbf{83.01} \\
\bottomrule
\end{tabular}%
}
\end{table*}

\begin{table*}[t]
\centering
\caption{Complete per-dimension ablation results for text-to-video (T2V) generation on VBench.}
\label{tab:app-full-vbench-t2v-ablation}

\resizebox{\textwidth}{!}{%
\begin{tabular}{llcccccccccc}
\toprule
\multirow{2}{*}{\textbf{Backbone}}
& \multirow{2}{*}{\textbf{Variant}}
& \textbf{Aesthetic} & \textbf{Appearance} & \textbf{Background}
& \multirow{2}{*}{\textbf{Color}}
& \textbf{Dynamic} & \textbf{Human} & \textbf{Imaging} & \textbf{Motion} & \textbf{Multiple} & \textbf{Object} \\
& & \textbf{Quality} & \textbf{Style} & \textbf{Consistency} & & \textbf{Degree} & \textbf{Action} & \textbf{Quality} & \textbf{Smoothness} & \textbf{Objects} & \textbf{Class} \\
\midrule

\rowcolor{gray!15}
\multirow{7}{*}{\textbf{Wan2.1-1.3B}}
& \textbf{TrajCFG (Full)} & \textbf{65.48} & \textbf{22.45} & 96.90 & 86.82 & \textbf{70.83} & 93.75 & 64.71 & 98.12 & \textbf{75.68} & \textbf{89.14} \\
& \textbf{Time-only} & 64.49 & 22.08 & \underline{97.14} & 85.06 & \underline{69.44} & \textbf{95.00} & \textbf{65.02} & \underline{98.16} & 72.26 & 86.55 \\
& \textbf{Sample-wise} & 64.17 & 22.13 & 97.02 & 86.77 & \underline{69.44} & \textbf{95.00} & 64.90 & 98.15 & 71.04 & 87.10 \\
& \textbf{Shuffled allocation} & 64.12 & 22.17 & 96.99 & 84.68 & \underline{69.44} & \underline{94.00} & 64.63 & 98.15 & 72.26 & \underline{87.74} \\
& \textbf{Reversed allocation} & 64.01 & 22.28 & 96.98 & 86.67 & \underline{69.44} & \underline{94.00} & 64.32 & \underline{98.16} & 69.59 & 85.21 \\
& \textbf{Adjacent-only $(1,0)$} & \underline{64.79} & \underline{22.31} & \textbf{97.18} & \underline{87.17} & 68.06 & \underline{94.00} & \underline{64.91} & 98.10 & \underline{75.08} & 87.34 \\
& \textbf{Stride-2-only $(0,1)$} & 64.71 & 22.22 & 97.05 & \textbf{87.57} & \underline{69.44} & \underline{94.00} & 64.40 & \textbf{98.18} & 73.09 & 86.16 \\
\midrule

\rowcolor{gray!15}
\multirow{7}{*}{\textbf{Wan2.2-5B}}
& \textbf{TrajCFG (Full)} & \textbf{65.49} & \textbf{22.43} & \textbf{97.53} & \underline{91.36} & \textbf{68.06} & \textbf{95.00} & \underline{67.26} & 98.45 & \textbf{73.86} & 85.68 \\
& \textbf{Time-only} & 65.35 & \underline{22.37} & 97.43 & 91.32 & \underline{66.67} & \underline{94.00} & 67.19 & 98.40 & \underline{72.26} & \underline{85.71} \\
& \textbf{Sample-wise} & \underline{65.47} & 22.22 & \underline{97.52} & 91.30 & 65.28 & 93.00 & 67.13 & 98.28 & 69.36 & 85.69 \\
& \textbf{Shuffled allocation} & 65.18 & 22.23 & 97.49 & 90.22 & \underline{66.67} & \textbf{95.00} & 66.88 & 98.18 & 71.49 & 85.48 \\
& \textbf{Reversed allocation} & 65.23 & 22.22 & 97.47 & 86.77 & \underline{66.67} & 92.00 & 66.59 & 98.39 & 71.11 & 85.18 \\
& \textbf{Adjacent-only $(1,0)$} & 65.44 & 22.31 & 97.41 & \textbf{93.26} & \textbf{68.06} & \underline{94.00} & \textbf{67.56} & \textbf{98.50} & 71.95 & 85.33 \\
& \textbf{Stride-2-only $(0,1)$} & 65.15 & 22.21 & 97.51 & 87.81 & \underline{66.67} & 93.00 & 66.89 & \underline{98.46} & 72.03 & \textbf{85.92} \\
\bottomrule
\end{tabular}%
}

\vspace{0.5em}

\resizebox{\textwidth}{!}{%
\begin{tabular}{llcccccc|cc|c}
\toprule
\multirow{2}{*}{\textbf{Backbone}}
& \multirow{2}{*}{\textbf{Variant}}
& \textbf{Overall}
& \multirow{2}{*}{\textbf{Scene}}
& \textbf{Spatial} & \textbf{Subject} & \textbf{Temporal} & \textbf{Temporal} & \textbf{Quality} & \textbf{Semantic}
& \multirow{2}{*}{\textbf{Total}} \\
& & \textbf{Consistency} & & \textbf{Relationship} & \textbf{Consistency} & \textbf{Flickering} & \textbf{Style} & \textbf{Score} & \textbf{Score} & \\
\midrule

\rowcolor{gray!15}
\multirow{7}{*}{\textbf{Wan2.1-1.3B}}
& \textbf{TrajCFG (Full)} & \textbf{25.61} & 39.64 & \textbf{72.79} & 93.13 & \underline{99.49} & \textbf{22.93} & \textbf{84.06} & \textbf{75.37} & \textbf{82.32} \\
& \textbf{Time-only} & 25.45 & 38.37 & 64.18 & \textbf{93.19} & \textbf{99.50} & 22.42 & \underline{83.94} & 73.17 & 81.78 \\
& \textbf{Sample-wise} & \underline{25.47} & \textbf{40.55} & 67.62 & \underline{93.16} & \textbf{99.50} & 22.52 & 83.83 & 74.02 & 81.87 \\
& \textbf{Shuffled allocation} & 25.46 & \underline{39.83} & 66.47 & 93.10 & \textbf{99.50} & 22.56 & 83.77 & 73.67 & 81.75 \\
& \textbf{Reversed allocation} & 25.44 & 39.31 & 64.03 & 93.10 & \textbf{99.50} & 22.38 & 83.70 & 72.96 & 81.55 \\
& \textbf{Adjacent-only $(1,0)$} & 25.41 & 39.03 & \underline{68.69} & 93.07 & \underline{99.49} & 22.58 & 83.81 & \underline{74.41} & \underline{81.93} \\
& \textbf{Stride-2-only $(0,1)$} & 25.44 & 37.79 & 63.98 & \textbf{93.19} & \underline{99.49} & \underline{22.59} & 83.86 & 73.39 & 81.77 \\
\midrule

\rowcolor{gray!15}
\multirow{7}{*}{\textbf{Wan2.2-5B}}
& \textbf{TrajCFG (Full)} & \textbf{25.84} & 49.49 & \textbf{77.19} & 94.46 & \underline{99.57} & 23.41 & \underline{84.82} & \textbf{77.45} & \textbf{83.35} \\
& \textbf{Time-only} & 25.74 & \underline{50.03} & 74.94 & 94.33 & \textbf{99.59} & 23.42 & 84.62 & \underline{76.94} & 83.08 \\
& \textbf{Sample-wise} & 25.67 & 49.78 & 74.51 & 94.35 & \textbf{99.59} & \underline{23.44} & 84.48 & 76.34 & 82.85 \\
& \textbf{Shuffled allocation} & \underline{25.75} & \textbf{50.09} & 74.56 & 94.29 & \textbf{99.59} & \underline{23.44} & 84.43 & 76.73 & 82.89 \\
& \textbf{Reversed allocation} & 25.71 & 49.65 & 74.61 & \underline{94.57} & \textbf{99.59} & 23.27 & 84.55 & 75.82 & 82.81 \\
& \textbf{Adjacent-only $(1,0)$} & 25.71 & 48.84 & \underline{75.24} & 94.46 & 99.56 & \textbf{23.47} & \textbf{84.85} & 76.93 & \underline{83.27} \\
& \textbf{Stride-2-only $(0,1)$} & 25.74 & 47.67 & 71.48 & \textbf{94.67} & \textbf{99.59} & 23.36 & 84.65 & 75.64 & 82.85 \\
\bottomrule
\end{tabular}%
}
\end{table*}

\begin{table*}[t]
\centering
\caption{Complete per-dimension results for text-to-video (T2V) generation on Dynamic-Bench.}
\label{tab:app-full-dynamic-bench}
\resizebox{\textwidth}{!}{%
\begin{tabular}{llcccccc}
\toprule
\multirow{2}{*}{\textbf{Backbone}}
& \multirow{2}{*}{\textbf{Method}}
& \textbf{Subject} & \textbf{Background} & \textbf{Motion} & \textbf{Aesthetic} & \textbf{Imaging} & \textbf{Overall} \\
& & \textbf{Consistency} & \textbf{Consistency} & \textbf{Smoothness} & \textbf{Quality} & \textbf{Quality} & \textbf{Consistency} \\
\midrule

\multirow{6}{*}{\textbf{Wan2.1-1.3B}}
& \textbf{Base}     & 85.54 & 91.42 & 96.40 & 45.17 & 56.21 & 22.36 \\
& \textbf{CFG-Zero} & 85.75 & 91.38 & 97.13 & 46.39 & 55.40 & 23.81 \\
& \textbf{CFG-Ctrl} & 86.80 & 92.47 & 97.28 & 44.40 & 56.27 & \textbf{24.18} \\
& \textbf{PAG}      & \underline{87.27} & 92.55 & 97.15 & 46.23 & \underline{56.40} & 23.89 \\
& \textbf{SRVS}     & \underline{87.27} & \underline{92.65} & \underline{97.38} & \underline{47.07} & 56.39 & 24.15 \\
\rowcolor{gray!15}
& \textbf{TrajCFG}  & \textbf{87.83} & \textbf{92.70} & \textbf{97.53} & \textbf{47.31} & \textbf{56.42} & \underline{24.17} \\
\midrule

\multirow{6}{*}{\textbf{Wan2.2-5B}}
& \textbf{Base}     & 86.50 & 92.84 & 96.58 & 45.82 & 58.95 & 24.00 \\
& \textbf{CFG-Zero} & 87.78 & 93.23 & 96.68 & 45.42 & \underline{60.35} & 24.19 \\
& \textbf{CFG-Ctrl} & 87.73 & 93.16 & \underline{96.95} & \underline{46.40} & 56.16 & 24.09 \\
& \textbf{PAG}      & 87.82 & \underline{93.40} & 96.91 & 43.23 & 60.29 & 24.11 \\
& \textbf{SRVS}     & \underline{88.12} & 93.39 & 96.87 & 46.13 & 59.88 & \underline{24.30} \\
\rowcolor{gray!15}
& \textbf{TrajCFG}  & \textbf{88.18} & \textbf{93.44} & \textbf{97.02} & \textbf{46.41} & \textbf{60.39} & \textbf{24.33} \\
\midrule

\multirow{6}{*}{\textbf{Wan2.2-14B}}
& \textbf{Base}     & 89.29 & 92.59 & 98.14 & 50.62 & 60.73 & 23.47 \\
& \textbf{CFG-Zero} & 88.97 & 92.44 & 98.02 & 50.69 & 61.14 & 25.18 \\
& \textbf{CFG-Ctrl} & \underline{90.13} & 92.71 & \underline{98.38} & 50.88 & 60.12 & 24.79 \\
& \textbf{PAG}      & 89.17 & 92.66 & 98.11 & \underline{50.98} & 60.97 & 24.90 \\
& \textbf{SRVS}     & 89.93 & \underline{92.82} & 98.26 & 50.63 & \underline{62.29} & \underline{25.24} \\
\rowcolor{gray!15}
& \textbf{TrajCFG}  & \textbf{90.74} & \textbf{93.12} & \textbf{98.43} & \textbf{51.24} & \textbf{62.80} & \textbf{25.31} \\
\bottomrule
\end{tabular}%
}
\end{table*}

\begin{table*}[t]
\centering
\caption{Complete per-dimension results for image-to-video (I2V) generation on VBench-I2V.}
\label{tab:app-full-vbench-i2v}
\resizebox{\textwidth}{!}{%
\begin{tabular}{llcccccccc}
\toprule
\multirow{2}{*}{\textbf{Backbone}}
& \multirow{2}{*}{\textbf{Method}}
& \textbf{I2V} & \textbf{I2V} & \textbf{Subject} & \textbf{Background}
& \textbf{Motion} & \textbf{Aesthetic} & \textbf{Imaging} & \textbf{Overall} \\
& & \textbf{Subject} & \textbf{Background} & \textbf{Consistency}
& \textbf{Consistency} & \textbf{Smoothness} & \textbf{Quality}
& \textbf{Quality} & \textbf{Consistency} \\
\midrule

\multirow{6}{*}{\textbf{Cosmos-Predict2.5-2B}}
& \textbf{Base}     & 98.18 & 98.74 & 96.00 & 96.18 & 99.09 & 62.23 & 71.51 & 23.48 \\
& \textbf{CFG-Zero} & \underline{98.26} & \underline{98.79} & \underline{96.39} & \underline{96.42} & \textbf{99.14} & \underline{62.46} & \underline{71.67} & \underline{23.50} \\
& \textbf{CFG-Ctrl} & 97.83 & 98.43 & 95.37 & 95.53 & 98.97 & 62.21 & 71.55 & 23.36 \\
& \textbf{PAG}      & 98.18 & 98.74 & 95.95 & 96.16 & 99.10 & 62.25 & 71.49 & 23.44 \\
& \textbf{SRVS}     & 97.77 & 98.47 & 94.68 & 96.07 & 99.11 & 62.27 & 71.48 & 23.42 \\
\rowcolor{gray!15}
& \textbf{TrajCFG}  & \textbf{98.29} & \textbf{98.81} & \textbf{96.51} & \textbf{96.49} & \underline{99.13} & \textbf{62.59} & \textbf{71.73} & \textbf{23.53} \\
\midrule

\multirow{6}{*}{\textbf{Wan2.2-5B}}
& \textbf{Base}     & 98.51 & 98.97 & 97.62 & 97.48 & 99.02 & 63.68 & 71.30 & 22.91 \\
& \textbf{CFG-Zero} & 97.95 & 98.55 & 96.22 & 96.54 & 98.92 & 62.46 & \underline{71.88} & 23.16 \\
& \textbf{CFG-Ctrl} & 98.04 & 98.69 & 96.33 & 96.24 & \underline{99.10} & 61.05 & 71.68 & 23.19 \\
& \textbf{PAG}      & 98.47 & 98.94 & 97.50 & 97.37 & 99.02 & 63.54 & 71.20 & 23.03 \\
& \textbf{SRVS}     & \underline{98.60} & \underline{99.02} & \underline{97.63} & \underline{97.65} & 99.09 & \underline{63.91} & \textbf{71.93} & \underline{23.22} \\
\rowcolor{gray!15}
& \textbf{TrajCFG}  & \textbf{98.78} & \textbf{99.16} & \textbf{98.33} & \textbf{98.04} & \textbf{99.32} & \textbf{64.36} & \textbf{71.93} & \textbf{23.24} \\
\midrule

\multirow{6}{*}{\textbf{Wan2.2-14B}}
& \textbf{Base}     & 98.28 & 98.74 & 94.74 & 95.04 & 98.77 & \underline{61.79} & 71.94 & 23.29 \\
& \textbf{CFG-Zero} & 97.35 & 98.03 & 97.07 & 96.48 & 98.69 & 60.24 & 71.06 & \underline{23.32} \\
& \textbf{CFG-Ctrl} & 98.24 & 98.72 & 96.59 & 96.02 & \underline{98.79} & 61.28 & \underline{71.98} & 23.21 \\
& \textbf{PAG}      & 97.90 & 98.49 & 96.98 & 96.42 & 98.76 & 61.78 & 71.91 & 23.24 \\
& \textbf{SRVS}     & \underline{98.36} & \underline{98.78} & \underline{97.10} & \underline{96.62} & \underline{98.79} & \underline{61.79} & 71.89 & 23.29 \\
\rowcolor{gray!15}
& \textbf{TrajCFG}  & \textbf{98.42} & \textbf{98.84} & \textbf{97.55} & \textbf{96.84} & \textbf{98.90} & \textbf{62.02} & \textbf{72.02} & \textbf{23.36} \\
\bottomrule
\end{tabular}%
}
\end{table*}

\begin{table*}[t]
\centering
\caption{Complete per-dimension results for image-to-video (I2V) generation on PAI-Bench-G.}
\label{tab:app-full-pai-bench-g}
\resizebox{\textwidth}{!}{%
\begin{tabular}{llcccccccc}
\toprule
\multirow{2}{*}{\textbf{Backbone}}
& \multirow{2}{*}{\textbf{Method}}
& \textbf{I2V} & \textbf{I2V} & \textbf{Subject} & \textbf{Background} & \textbf{Motion} & \textbf{Aesthetic} & \textbf{Imaging} & \textbf{Overall} \\
& & \textbf{Subject} & \textbf{Background} & \textbf{Consistency} & \textbf{Consistency} & \textbf{Smoothness} & \textbf{Quality} & \textbf{Quality} & \textbf{Consistency} \\
\midrule

\multirow{6}{*}{\textbf{Cosmos-Predict2.5-2B}}
& \textbf{Base}     & 96.27 & 97.25 & 92.43 & 94.17 & 99.09 & 52.46 & 70.18 & \underline{21.18} \\
& \textbf{CFG-Zero} & \underline{96.33} & \underline{97.32} & \underline{92.70} & \underline{94.38} & \underline{99.13} & 52.83 & 70.38 & \textbf{21.19} \\
& \textbf{CFG-Ctrl} & 95.33 & 96.36 & 91.23 & 93.54 & 98.37 & 51.59 & 70.27 & 20.94 \\
& \textbf{PAG}      & 96.21 & 97.21 & 92.14 & 94.08 & 99.08 & \textbf{52.87} & 70.36 & 21.17 \\
& \textbf{SRVS}     & 95.40 & 96.65 & 91.67 & 93.71 & 99.10 & \underline{52.86} & \underline{70.43} & 21.15 \\
\rowcolor{gray!15}
& \textbf{TrajCFG}  & \textbf{96.39} & \textbf{97.38} & \textbf{92.77} & \textbf{94.44} & \textbf{99.14} & 52.62 & \textbf{70.46} & 21.15 \\
\midrule

\multirow{6}{*}{\textbf{Wan2.2-5B}}
& \textbf{Base}     & 96.47 & 97.80 & 92.72 & \underline{94.57} & 98.72 & 51.24 & 67.88 & 21.26 \\
& \textbf{CFG-Zero} & 96.17 & 97.55 & 92.64 & 94.47 & 98.72 & 51.58 & 68.41 & 21.27 \\
& \textbf{CFG-Ctrl} & \underline{96.53} & \underline{97.92} & \underline{92.80} & 94.21 & \underline{98.97} & 50.81 & 68.80 & 21.06 \\
& \textbf{PAG}      & 96.27 & 97.63 & 92.70 & 94.46 & 98.69 & \underline{51.65} & \underline{68.86} & \underline{21.33} \\
& \textbf{SRVS}     & 96.11 & 97.38 & 92.10 & 94.39 & 98.50 & \textbf{51.82} & \textbf{68.96} & \underline{21.33} \\
\rowcolor{gray!15}
& \textbf{TrajCFG}  & \textbf{96.95} & \textbf{98.13} & \textbf{94.01} & \textbf{95.23} & \textbf{98.99} & 51.45 & 67.93 & \textbf{21.35} \\
\midrule

\multirow{6}{*}{\textbf{Wan2.2-14B}}
& \textbf{Base}     & 96.12 & 97.26 & 91.81 & \underline{92.32} & 98.50 & \underline{51.11} & 66.60 & 21.36 \\
& \textbf{CFG-Zero} & 95.84 & 97.04 & 91.05 & 91.81 & 98.48 & 51.09 & 65.95 & 21.41 \\
& \textbf{CFG-Ctrl} & 95.80 & 97.04 & \underline{91.89} & 92.02 & 98.52 & 50.51 & 65.84 & 21.30 \\
& \textbf{PAG}      & 96.08 & 97.24 & 91.72 & 92.30 & 98.53 & 50.95 & 66.42 & \underline{21.43} \\
& \textbf{SRVS}     & \underline{96.17} & \underline{97.29} & 91.71 & 92.26 & \underline{98.61} & 51.09 & \underline{66.62} & \underline{21.43} \\
\rowcolor{gray!15}
& \textbf{TrajCFG}  & \textbf{96.37} & \textbf{97.51} & \textbf{92.77} & \textbf{92.72} & \textbf{98.65} & \textbf{51.13} & \textbf{66.67} & \textbf{21.47} \\
\bottomrule
\end{tabular}%
}
\end{table*}

\begin{table*}[t]
\centering
\caption{Complete per-dimension results for text-to-audio-video (T2AV) generation on VABench. Lower is better for Desync.}
\label{tab:app-full-vabench-t2av}
\resizebox{\textwidth}{!}{%
\begin{tabular}{ll|ccccccc|cccccc}
\toprule
\multirow{2}{*}{\textbf{Backbone}}
& \multirow{2}{*}{\textbf{Method}}
& \textbf{Speech} & \textbf{Audio} & \textbf{T--V} & \textbf{T--A}
& \textbf{A--V} & \textbf{Lip--Sync}
& \multirow{2}{*}{\textbf{Desync}$\downarrow$}
& \multirow{2}{*}{\textbf{Alignment}}
& \multirow{2}{*}{\textbf{Expressiveness}}
& \textbf{Visual} & \textbf{Audio} & \textbf{Audio} & \textbf{Visual} \\
& & \textbf{Q\&N} & \textbf{Aes.} & \textbf{Align.} & \textbf{Align.}
& \textbf{Align.} & & & & & \textbf{Realism} & \textbf{Realism}
& \textbf{QA} & \textbf{QA} \\
\midrule

\multirow{2}{*}{\textbf{LTX-2.3}}
& \textbf{Base}
& 2.698 & 3.345 & 0.1928
& 0.3118 & 0.2490 & \textbf{2.752}
& 0.5911 & 4.470 & 4.299
& 4.615 & 4.032 & 0.7049
& 0.6616 \\
& \textbf{TrajCFG}
& \textbf{2.763} & \textbf{3.351} & \textbf{0.1963}
& \textbf{0.3202} & \textbf{0.2567} & 2.555
& \textbf{0.5685} & \textbf{4.479} & \textbf{4.321}
& \textbf{4.638} & \textbf{4.066} & \textbf{0.7178}
& \textbf{0.6730} \\
\midrule

\multirow{2}{*}{\textbf{Ovi}}
& \textbf{Base}
& 3.181 & 3.282 & 0.1968
& 0.3288 & 0.1564 & 4.420
& 0.7659 & \textbf{4.189} & 4.138
& 4.522 & 4.063 & 0.5251
& 0.5948 \\
& \textbf{TrajCFG}
& \textbf{3.463} & \textbf{3.364} & \textbf{0.1977}
& \textbf{0.3316} & \textbf{0.1654} & \textbf{4.780}
& \textbf{0.7080} & 4.179 & \textbf{4.141}
& \textbf{4.547} & \textbf{4.074} & \textbf{0.5257}
& \textbf{0.5953} \\
\bottomrule
\end{tabular}%
}
\end{table*}

\begin{table*}[t]
\centering
\caption{Complete per-dimension results for text-to-audio-video (T2AV) generation on T2AV-Compass. Lower is better for DS.}
\label{tab:app-full-t2av-compass}
\resizebox{\textwidth}{!}{%
\begin{tabular}{ll|cc|cc|ccccc}
\toprule
\multirow{2}{*}{\textbf{Backbone}}
& \multirow{2}{*}{\textbf{Method}}
& \multicolumn{2}{c|}{\textbf{Video Quality}}
& \multicolumn{2}{c|}{\textbf{Audio Quality}}
& \multicolumn{5}{c}{\textbf{Cross-modal Alignment}} \\
\cmidrule(lr){3-4}\cmidrule(lr){5-6}\cmidrule(lr){7-11}
& & \textbf{VT}$\uparrow$ & \textbf{VA}$\uparrow$
& \textbf{AA}$\uparrow$ & \textbf{SQ}$\uparrow$
& \textbf{A--V}$\uparrow$ & \textbf{T--A}$\uparrow$
& \textbf{T--V}$\uparrow$ & \textbf{DS}$\downarrow$
& \textbf{LS}$\uparrow$ \\
\midrule

\multirow{2}{*}{\textbf{LTX-2.3}}
& \textbf{Base}
& 7.295 & 4.921
& 6.709 & 1.562
& 0.2469 & 0.1722
& \textbf{0.0432} & 0.7140
& 2.484 \\
& \textbf{TrajCFG}
& \textbf{7.510} & \textbf{4.924}
& \textbf{6.738} & \textbf{1.578}
& \textbf{0.2481} & \textbf{0.1735}
& 0.0423 & \textbf{0.7016}
& \textbf{2.820} \\
\midrule

\multirow{2}{*}{\textbf{Ovi}}
& \textbf{Base}
& 6.714 & 4.516
& 6.676 & 1.566
& 0.1596 & 0.0897
& 0.0297 & 0.9624
& 1.939 \\
& \textbf{TrajCFG}
& \textbf{7.199} & \textbf{4.555}
& \textbf{6.733} & \textbf{1.665}
& \textbf{0.1622} & \textbf{0.0922}
& \textbf{0.0316} & \textbf{0.9356}
& \textbf{2.449} \\
\bottomrule
\end{tabular}%
}
\end{table*}

\clearpage
\bibliographystyle{iclr2027_conference}
\bibliography{iclr2027_conference}

\end{document}